\def\arxivbuild{}
\PassOptionsToPackage{table}{xcolor}
\documentclass{article}
\usepackage{iclr2027_conference,times}

\usepackage{amsmath,amsfonts,bm}

\def\eqref#1{equation~\ref{#1}}

\def\1{\bm{1}}

\DeclareMathAlphabet{\mathsfit}{\encodingdefault}{\sfdefault}{m}{sl}
\SetMathAlphabet{\mathsfit}{bold}{\encodingdefault}{\sfdefault}{bx}{n}

\usepackage{amsmath}
\usepackage{amssymb}
\usepackage{array}
\usepackage{booktabs}
\usepackage{float}
\usepackage{graphicx}
\usepackage{microtype}
\usepackage{multirow}
\usepackage{xcolor}
\usepackage{textcomp}
\usepackage[normalem]{ulem}
\usepackage[T1,OT1]{fontenc} %
\usepackage{hyperref}
\usepackage{url}
\usepackage[caption=false,font=small]{subfig}
\usepackage[normalem]{ulem}
\graphicspath{{assets/}}
\definecolor{bandgray}{gray}{0.92}
\definecolor{ourrow}{RGB}{235,235,250}
\definecolor{linkcolor}{RGB}{18,88,138}
\hypersetup{hidelinks,pdftitle={Beyond the Timeline: Augmenting Long-Video Memory with Grounded Entity Biographies}}

\newif\ifarxiv
\ifdefined\arxivbuild\arxivtrue\else\arxivfalse\fi
\newcommand{\name}{GEB}
\newcommand{\fullname}{Grounded Entity Biographies}

\title{Beyond the Timeline: Augmenting Long-Video Memory with \fullname{}}

\ifarxiv
  \iclrfinalcopy
  \author{Hui Ren\textsuperscript{1} \thanks{Work done during an internship at Amazon.},\, Lei Fan\textsuperscript{2}, Henry Pao\textsuperscript{2}, Han Guo\textsuperscript{2}, \\
  \bfseries Zeeshan Zia\textsuperscript{2}, Ying Chen\textsuperscript{2}, Alexander Schwing\textsuperscript{1}, Gang Hua\textsuperscript{2} \\[4pt]
  {\normalfont \textsuperscript{1}University of Illinois Urbana-Champaign \qquad \textsuperscript{2}Amazon.com, Inc.}
}
\newcommand{\projectpage}{https://geb-video.github.io}
\newcommand{\pdfauthors}{Hui Ren, Lei Fan, Henry Pao, Han Guo, Zeeshan Zia, Ying Chen, Alexander Schwing, Gang Hua}

  \hypersetup{pdfauthor={\pdfauthors}}
\else
  \author{Anonymous authors\\
  Paper under double-blind review}
\fi
\begin{document}

\maketitle
\ifarxiv\lhead{}\vspace{-14pt}\centerline{\href{\projectpage}{\textcolor{linkcolor}{\textbf{Project page}}}}\vspace{6pt}\fi

\begin{abstract}
Answering questions about long videos often requires connecting events involving the same objects across hours or days. 
Chronological descriptions and text-derived entities can leave physical identity unresolved: different objects may share a description, while observations of the same object remain disconnected across events.
Retrieving relevant events therefore does not necessarily recover the ``biography'' of the particular entity a question concerns. 
To address this, we introduce {\emph{\fullname}} (\name), a long-video memory framework that groups visually grounded observations of the same physical instance across clips into retrievable biographies while preserving the context of each moment.
During question answering, the biography is retrieved alongside episodic evidence, allowing the model to follow an entity through events using identity links established during memory construction.
Evaluations across four benchmarks, including day-long and week-long recordings, demonstrate improvements over prior memory frameworks in both multiple-choice and open-ended question answering. 
On EgoLifeQA, \name{} achieves $72.0\%$ accuracy, $4.4$ percentage points above the best published result. 
Ablations show that grounded identity association and biography reading both contribute to the gains, which additional descriptions alone do not fully recover.

\end{abstract}

\section{Introduction}
\label{sec:introduction}

History can be organized around events or around the subjects who took part. A chronicle follows events through time; a biography follows a subject through those events. Long-video memory needs both perspectives: it must recover what happened at a particular moment and connect what happened to the same person or object across hours or days. The latter requires deciding which scattered observations concern the same physical entity. Without this correspondence, a detailed record of moments leaves the biography of an entity incomplete. 

\begin{figure}[t]
  \centering
  \includegraphics[width=1\textwidth]{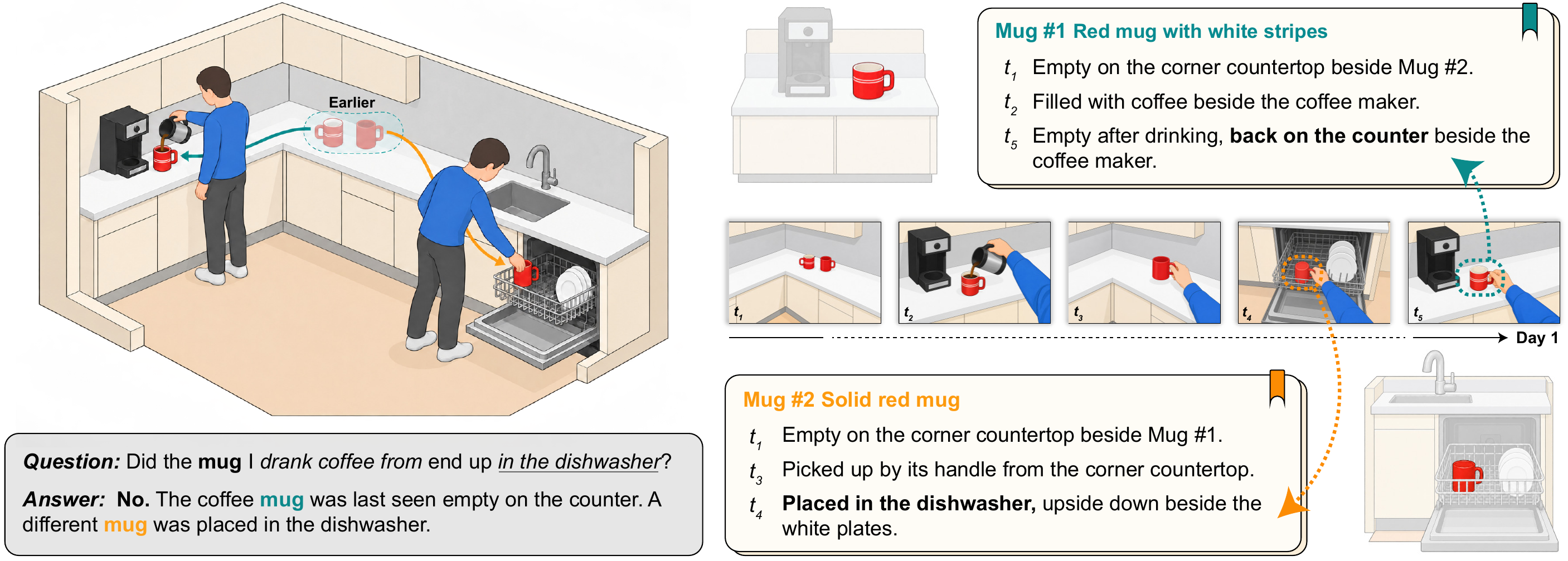}
  \caption{\textbf{Moments record events; persistent entities connect them into biographies.} To answer whether the mug used for coffee ended up in the dishwasher, a memory must know which physical mug took part in each event. A descriptive memory may retrieve ``coffee is poured into a red mug'' and ``a red mug is placed in the dishwasher'' yet cannot tell whether the two mugs are the same.}
  \label{fig:teaser}
  \vspace{-2mm}
\end{figure}

Consider the question in Figure~\ref{fig:teaser}: \emph{Did the mug I drank coffee from end up in the dishwasher?} The striped red mug was used for coffee and later seen empty on the counter; a different, solid red mug was placed in the dishwasher. A descriptive memory may retrieve both ``coffee is poured into a red mug'' and ``a red mug is placed in the dishwasher.''
Both descriptions can be accurate, yet their shared wording does not establish that the events involve the same instance. Retrieving relevant events is therefore insufficient without resolving whose biography they belong to. 

Memory frameworks make long recordings searchable through temporal descriptions and semantic relations~\citep{wang2024videoagent,yeo2025worldmm}, with recent structured memory frameworks also consolidating entity mentions into cross-time narratives~\citep{li2026magic}. When correspondence is derived from language, two related limitations remain. First, descriptions can conflate or fragment the biographies of physical objects: different objects may share a name, while one object may be described differently as its state, location, or activity changes. Second, retrieving an observation may not reveal the other encounters of that particular instance, especially across recording sessions, where temporal proximity and continuous tracking no longer connect observations. Without identity established in the memory itself, retrieval and reasoning must reconstruct which moments belong together from the evidence available for each question. 

We introduce \emph{\fullname} (\name), a long-video memory framework that organizes observations into biographies of inferred physical instances while preserving the context and visual evidence of each encounter (Figure~\ref{fig:method}). \name{} grounds each observation in views of the instance and its surrounding activity, then associates observations across clips using visual and contextual evidence, and rejects a match when the video shows two different physical instances. At question time, a retrieved observation becomes an entry point: retrieval follows same-instance edges to the other observations of the same inferred instance and can reach their episodic context and source evidence. Further, the biography excerpt presented to the controller also lists the observations not yet inspected, giving the controller concrete targets for further search.

We evaluate \name{} on day-long and week-long recordings, covering multiple-choice and open-ended question answering~\citep{yang2025egolife,tian2025egor1,chen2026mmlifelong}. On EgoLifeQA, \name{} achieves $72.0\%$ accuracy versus $67.6\%$ reported by MAGIC-Video, the strongest published memory framework, while using the same retrieval controller, the same answer model, and the same retrieval limits. The fraction of questions whose evidence window reaches the answering context rises from $37.6\%$ to $58.9\%$, indicating better access to relevant moments. Ablations show: indexing descriptions without physical-instance association recovers only part of the gain; removing the same-instance edges, the edges to episodic context, or the biography text each reduces accuracy.

Our contributions are summarized as follows:
\begin{itemize}
  \item \textbf{Grounded entity biographies:} a memory representation and construction approach that links visually grounded observations into persistent entity biographies while retaining their event context and supporting evidence.
  \item \textbf{Retrieval and reading through identity:} a mechanism that connects episodes through shared physical entities and presents biography excerpts that support reasoning and guide further search.
  \item \textbf{Empirical validation and analysis:} improvements on long-video question answering under matched reasoning components, with evidence-access diagnostics and ablations examining the roles of grounding, association, and biography reading.
\end{itemize}

\section{Related Work}
\label{sec:related}

\noindent\textbf{Long-video memory and agentic retrieval.}
Long-video models extend the amount of visual context they can process through memory compression and hierarchical token reduction~\citep{song2024moviechat,li2025videochatflash}. Retrieval-based approaches access selected evidence on demand: VideoAgent iteratively gathers information with visual tools~\citep{wang2024videoagent}, while WorldMM coordinates retrieval from episodic, semantic, and visual memories~\citep{yeo2025worldmm}. Ego-R1 learns to compose tool calls for long-horizon reasoning~\citep{tian2025egor1}, and ReMA recursively manages a multimodal belief state~\citep{chen2026mmlifelong}. \name{} complements these retrieval strategies by connecting a matched observation to other events involving the same physical instance.

\noindent\textbf{Entity memory.}
Entity-centered memory frameworks retain information about recurring people and objects. VideoAgent tracks and re-identifies objects within a video and keeps an object-occurrence database~\citep{fan2024videoagentmem}; AMEGO links the interaction tracklets of one object and records where it is used~\citep{goletto2024amego}. M3-Agent gives the people in a video face and voice identities beside episodic and semantic text~\citep{long2026m3}, and Embodied VideoAgent maintains persistent objects from egocentric video with depth and pose sensing~\citep{fan2025embodied}. 
\name{} organizes observations of the same physical instance into biographies while retaining event contexts. These biographies support retrieval across events and guide further search through references to other recorded appearances.

\noindent\textbf{Structured memory and relational retrieval.}
Graph-based retrieval connects evidence through extracted entities and relations. GraphRAG organizes document collections through entity graphs and community summaries~\citep{edge2024graphrag}; HippoRAG and HippoRAG~2 support associative retrieval through knowledge graphs and links to source passages~\citep{gutierrez2024hipporag,gutierrez2025hipporag2}. For video, EGAgent and EgoGraph construct temporal entity graphs from transcripts and scene descriptions~\citep{rege2026egagent,sun2026egograph}. MAGIC-Video builds a multimodal memory graph over its captions, using entity nodes for names extracted by a language model  and consolidated across time, augmented by topic and event chains~\citep{li2026magic}. 
Language-derived entity links can merge distinct physical instances
or split one instance across names. \name{} associates observations
using visual and contextual evidence, with separation in shared
frames vetoing a match.

\section{\fullname}
\label{sec:method}
\fullname{} (\name) connects persistent entity biographies to episodic memory. We define this representation (Sections~\ref{sec:method-biographies}--\ref{sec:method-schema}), then describe how grounded association constructs biographies and retrieval follows them across events (Sections~\ref{sec:method-build}--\ref{sec:method-infer}).

\begin{figure}[t]
  \centering
  \includegraphics[width=\textwidth]{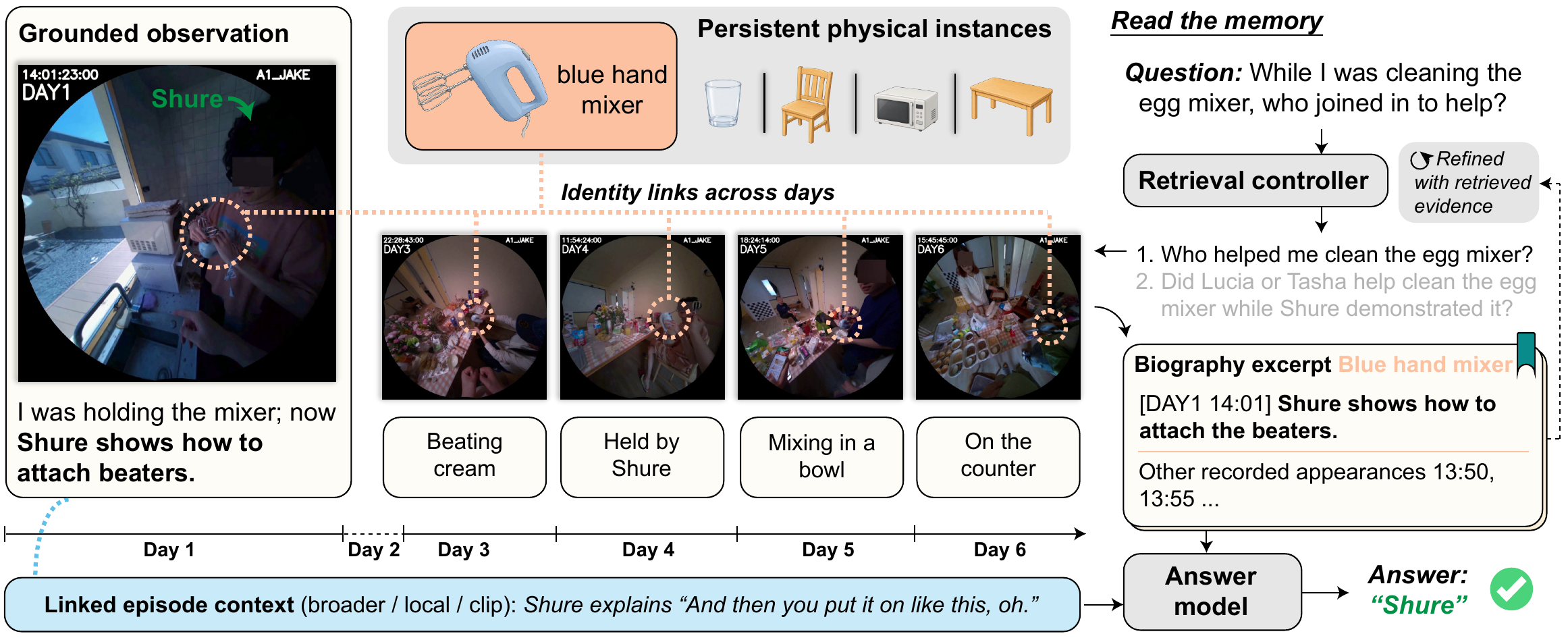}
  \caption{\textbf{Overview of \fullname{} (\name).} Visually grounded observations of each instance are associated across clips into a persistent biography; here the blue hand mixer is linked across days, each observation keeping its description, its source frames and its episode context. At question time a controller retrieves biography excerpts together with episodic and visual evidence, and the answer model identifies Shure from them.}
  \label{fig:method}
  \vspace{-2mm}
\end{figure}

\subsection{Representing an Entity Biography}
\label{sec:method-biographies}

A grounded entity biography is a temporally ordered record of encounters with one physical entity. Each encounter preserves the visible subject, its event context, and the supporting evidence.

We partition a given recording into $T$ short video clips. Each $c_t$ denotes one clip containing a sequence of frames, and $\mathcal{C}=\{c_t\}_{t=1}^{T}$ denotes the collection of clips. An \emph{observation} $o_i$ records one tracked subject within one clip; $\mathcal{O}$ denotes all such observations. Each record retains the visible time span $\tau_i$ of the subject in the recording, tracked image regions, a description of its state and interactions, and references to its source clip and supporting context. The context used to describe an encounter may extend beyond its visible span. For example, the mixer being handled in one clip and resting on a counter in another constitute two observations, even if they depict the same mixer (Figure~\ref{fig:method}).

Let $\hat z_i\in\{1,\ldots,K\}$ be the entity identifier assigned to observation $o_i$, where $K$ is the number of inferred physical entities. The biography of entity $k$ is
\begin{equation}
    \mathcal{B}_k
    =
    \operatorname{sort}_{\tau}
    \bigl(\{o_i\in\mathcal{O}:\hat z_i=k\}\bigr),
    \label{eq:biography}
\end{equation}
where $\operatorname{sort}_{\tau}$ orders observations by the start times of their visible spans. Membership expresses estimated correspondence to the same physical instance across clips; a shared category or name alone does not establish it. Gaps in this observed history do not imply that the entity was absent or inactive. Section~\ref{sec:method-build} discusses how $\hat z_i$ is computed.

\subsection{Connecting Biographies to Episodic Memory}
\label{sec:method-schema}

A biography connects encounters with the same subject, while an episode preserves the surrounding activity and other participants needed to interpret them. Identifying who helped with the mixer, for example, requires context about the people involved. In our memory, we therefore link each encounter to its biography, surrounding episode, and source video.

A \emph{persistent entity node} $u_k$ represents one inferred physical instance, such as the blue mixer across days, without requiring continuous visibility. \emph{Observation nodes} retain its individual encounters. \emph{Episode nodes} hold scene descriptions at multiple temporal scales, such as Shure's actions and dialogue while handling the mixer (Figure~\ref{fig:method}). \emph{Source-clip nodes} reference the original video. An episode and source clip may cover the same interval but supply different evidence: a contextual description and its supporting frames.

Formally, let $\mathcal{U}$ and $\mathcal{E}$ denote the sets of entity and episode nodes. Using the above notation for nodes of observations (${\cal O}$) and clip records (${\cal C}$), we represent the memory as
\begin{equation}
    \mathcal{G}=(\mathcal{N},\mathcal{R}),
    \qquad
    \mathcal{N}
    =
    \mathcal{U}\cup\mathcal{O}\cup\mathcal{E}\cup\mathcal{C}.
    \label{eq:memory-graph}
\end{equation}
Here, $\mathcal{N}$ is the complete node set and $\mathcal{R}$ is the set of all relation edges described next.

Three relation types connect observations to identity, context, and visual evidence. \emph{Entity membership} contributes an edge $(u_k,o_i)\in\mathcal{R}$ whenever observation $o_i$ is assigned to entity $k$, i.e., $\hat z_i=k$. \emph{Episode context} links each observation to its local episode. \emph{Visual provenance} links it to its source clip. Within episodic memory, \emph{temporal adjacency} connects successive episodes at the same scale, while \emph{containment} connects local episodes to coarser episodes that contain them, providing access to broader activity context.

\textbf{Persistent entities act as bridges across episodes.} For observations $o_i$ and $o_j$ assigned to entity $k$, membership and context links establish a path
\[
    e_a
    \;\leftrightarrow\;
    o_i
    \;\leftrightarrow\;
    u_k
    \;\leftrightarrow\;
    o_j
    \;\leftrightarrow\;
    e_b,
\]
where $e_a$ and $e_b$ contain the two observations. The shared identity connects encounters across days even when their descriptions differ; linked episodes preserve who interacted with it at each moment. The ablations of Section~\ref{sec:exp-ablation} distinguish \textsc{same-instance} edges (entity membership) from observation$\to$timeline edges (episode context and visual provenance).

\subsection{Writing the Memory}
\label{sec:method-build}

Writing a biography requires attributing an event to the correct visible subject and recognizing that subject when it reappears. \name{} separates these decisions: grounded descriptions preserve individual encounters, while cross-clip association determines which encounters share an entity. 

\noindent\textbf{Grounding encounters in their context.}
An open-vocabulary detector and a within-clip tracker group repeated detections into observations, providing multiple views of a subject within one encounter. To describe each encounter, a vision-language model jointly reads subject crops, scene frames, and available timestamped narration and dialogue. Crops reveal distinguishing details, scene frames show interactions involving the subject, and text supplies surrounding event context. The prompt instructs the model to establish the subject visually and use textual context only when it concerns that subject. This addresses a central attribution problem: an action described near an object need not involve that object.

\noindent\textbf{Associating encounters conservatively.}
For each new observation, we decide whether to append it to an existing biography or start a new one. A mistaken association creates false paths between events, so a match must have positive support and remain consistent with the retained evidence. 

For this, we process observations in temporal order, retrieving candidate matches among existing entities by embedding similarity. Let $\mathbf{v}_i$ be a normalized multimodal embedding of the new observation $o_i$; $s_{ij}=\mathbf{v}_i^\top\mathbf{v}_j$ measures its similarity to an earlier observation $o_j$. For an existing candidate entity $k$, the nonempty set $M_k$ contains a bounded number of its most recent assigned observations, used as comparison references.

Two thresholds, $\theta_{\mathrm{cons}}<\theta_{\mathrm{match}}$, impose complementary requirements. The new observation must be sufficiently similar to \emph{every} reference, guarding against inconsistent views being combined, and strongly match \emph{at least one}, requiring positive correspondence evidence. 
A third check uses visual separation: we compute bounding-box intersection-over-union (IoU) on frames shared by the two observations. We set $D_{ij}=1$ when the two observations share at least a minimum number of frames and their boxes fall below the IoU threshold in at least a prescribed fraction of those frames, and $D_{ij}=0$ otherwise (Appendix~\ref{app:implementation}).
A zero means that no separation evidence was found; it does not establish that the subjects are the same instance.

We assign $o_i$ to a candidate entity $k$ only if all three conditions hold:
\begin{equation}
    \min_{o_j\in M_k}s_{ij}\ge\theta_{\mathrm{cons}},
    \quad
    \max_{o_j\in M_k}s_{ij}\ge\theta_{\mathrm{match}},
    \quad
    D_{ij}=0\;\;\forall\,o_j\in M_k.
    \label{eq:association}
\end{equation}
Thus, matching one reference cannot compensate for contradicting another. If a reference and the new observation show two similar mixers apart in shared frames, that candidate is rejected regardless of embedding similarity.

If multiple candidates qualify, we choose the one with the highest mean similarity to its reference observations.
For the selected entity $k$, we set $\hat z_i=k$ and append $o_i$ to its reference set $M_k$, removing the oldest reference if the size limit is exceeded; if no candidate qualifies, $o_i$ starts a new entity.
Bounding $M_k$ limits comparison cost, while all assigned observations remain in $\mathcal{B}_k$. Threshold calibration and separation-test settings are given in Appendix~\ref{app:implementation}.

People follow the same observation and association process as objects. When participants are named, an additional stage assigns and consolidates identities using unambiguous person-and-day appearance profiles, abstaining when identifying features are shared (Appendix~\ref{app:implementation}).

\subsection{Reading the Memory}
\label{sec:method-infer}

A question may identify an entity through one encounter but concern another. Retrieval uses the matched encounter to enter its biography and recover surrounding evidence. A language-model controller~\citep{li2026magic} reads the question and accumulated evidence, then issues another search or passes deduplicated biographies, episode excerpts, and source frames to the answer model.

\noindent\textbf{Retrieving through identity and context.}
Observation descriptions and episode captions are indexed together, semantically and lexically; a query's initial matches come from this shared index and from visual matches to source clips. Relevance propagates from these matches through entity membership to other observations of the same inferred instance, and through context links to their surrounding episodes. We implement this propagation with Personalized PageRank~\citep{haveliwala2002topicsensitive} and combine its scores with query-text similarity for ranking. Entity nodes transmit relevance; the returned evidence consists of observations, episodes, and source clips.

These records compete within a shared retrieval budget. Limits on selected appearances, both overall and per entity, preserve room for episodic context and prevent a frequently observed subject from dominating. Relation weights and budget settings are specified in Appendix~\ref{app:implementation}. For timestamped questions, retrieval and rendering are restricted to memory records preceding the query time.

\noindent\textbf{Reading a biography and extending the search.}
For each entity, the biography excerpt contains its observations selected through the current retrieval round, ordered by time.
Each block retains the entity identifier, timestamps, and stored descriptions. Other recorded appearances that have not been selected are summarized by times and counts, giving the controller concrete targets for subsequent searches. The biography therefore provides both evidence about the subject and access to parts of its history that remain to be inspected.

Conservative association can leave one physical instance under multiple identifiers. When biographies that share a name are retrieved together, an accompanying note distinguishes pairs with visual evidence of separation from those whose identity remains unresolved. Different identifiers alone are not treated as proof of different objects. Linked episodic evidence also provides context for checking the account of an observation; the reading prompt gives the episode precedence when the two descriptions conflict.

\section{Experiments}
\label{sec:experiments}

We evaluate question answering, access to supporting moments, and the contributions of identity association, episodic connections, and biography reading.

\subsection{Experimental Setup}
\label{sec:exp-setup}

\noindent\textbf{Benchmarks.}
Our evaluation covers complementary demands of video memory. \emph{EgoLifeQA}~\citep{yang2025egolife} and \emph{Ego-R1-Bench}~\citep{tian2025egor1} test multiple-choice answering over approximately 52 hours of participant A1's week, using 500 and 50 questions, respectively. Only recordings preceding each question's timestamp are accessible. \emph{MM-Lifelong}~\citep{chen2026mmlifelong} tests open-ended answering over the same week (Test@Week) and a 23.6-hour gameplay stream (Test@Day), with the complete recording accessible. The same EgoLife memory supports both multiple-choice benchmarks and Test@Week without rebuilding. \emph{MultiHop-EgoQA}~\citep{chen2024multihop} tests questions requiring evidence from separate moments: it contains 1,080 questions over 360 three-minute Ego4D clips~\citep{grauman2022ego4d}. Its annotated evidence intervals allow us to evaluate whether retrieval reaches all required moments. Memory construction uses video without audio.

\noindent\textbf{Comparisons.}
We compare with general, long-video, and agentic video models, distinguishing published results from our runs in the tables. For comparisons among memory frameworks, MAGIC-Video~\citep{li2026magic}, WorldMM~\citep{yeo2025worldmm}, and \name{} use Qwen3.5-35B as controller and answer model. Their EgoLifeQA and Ego-R1-Bench baseline results are taken from \citet{li2026magic}. On MM-Lifelong and MultiHop-EgoQA, we run the released implementations. \name{} and MAGIC-Video share the episodic captions, topic and event summaries, and retrieval limits. WorldMM retains its own episodic, semantic, and visual stores. Appendix~\ref{app:implementation} specifies retrieval-unit, search-round and frame limits, and Appendix~\ref{app:protocol} the per-benchmark protocol, including WorldMM's per-store retrieval. No-memory reference rows evaluate the answer model with sampled video evidence. On MultiHop-EgoQA, this reference receives frames spanning the entire clip, whereas memory frameworks select evidence through retrieval.

\noindent\textbf{Evaluation.}
We report multiple-choice accuracy on EgoLifeQA and Ego-R1-Bench, and the benchmark's GPT-5-judged accuracy on MM-Lifelong. MultiHop-EgoQA evaluates answer quality and temporal grounding. We use its released scoring code and its grading prompt with an independent \texttt{gpt-oss-120b} judge. Grading uses the 724 questions with a reference answer. Ego-R1-Bench results are averaged over three seeds, and our memory-framework evaluations on MM-Lifelong and MultiHop-EgoQA each report the mean over three runs. Detailed scoring protocols appear in Appendix~\ref{app:protocol}. Uncertainty estimates are summarized with the main results below.

\subsection{Question Answering Results}
\label{sec:exp-main}

\name{} achieves the highest overall score among the compared systems on all four splits in Table~\ref{tab:main}.

\begin{table}[t]
\caption{\textbf{\name{} achieves the highest overall accuracy on each benchmark split shown.} Accuracy ($\%$). EL, ER, HI, RM, TM: EntityLog, EventRecall, HabitInsight, RelationMap, TaskMaster; Manual/Gemini: human-written/model-generated Ego-R1 questions. $^*$ on a model name: EgoLifeQA and Ego-R1 results from \citet{li2026magic}; $^*$ on a cell: \citet{chen2026mmlifelong}; $^\dagger$: \citet{yeo2025worldmm}. Other entries are our runs. GEB, MAGIC-Video, and WorldMM use the same Qwen3.5-35B controller and answer model. \textbf{Bold}/\underline{underline}: best/second-best per column. Full comparisons in Tables~\ref{tab:main-full} and~\ref{tab:mmlifelong-full}.}
\label{tab:main}
\centering
\small
\setlength{\tabcolsep}{2.4pt}
\begin{tabular}{@{}l@{\hspace{4pt}}cccccc@{\hspace{6pt}}ccc@{\hspace{7pt}}cc@{}}
\toprule
& \multicolumn{6}{c}{\bf EgoLifeQA $\uparrow$} & \multicolumn{3}{c}{\bf Ego-R1 $\uparrow$} & \multicolumn{2}{c}{\bf MM-Lifelong $\uparrow$} \\
\cmidrule(lr){2-7} \cmidrule(lr){8-10} \cmidrule(lr){11-12}
Model & EL & ER & HI & RM & TM & Avg. & Manual & Gemini & Avg. & Week & Day \\
\midrule
\rowcolor{bandgray} \multicolumn{12}{c}{\bf General MLLMs} \\
GPT-5$^\dagger$ & 47.2 & 42.1 & 47.5 & 53.6 & 55.6 & 48.6 & -- & -- & -- & 15.00$^*$ & 15.25$^*$ \\
Qwen3-VL-235B & 44.0 & 44.4 & 50.8 & 44.8 & 50.8 & 46.0 & 40.0 & 62.7 & 51.3 & 15.63$^*$ & 12.44$^*$ \\
Qwen3.5-35B & 43.2 & 46.8 & 49.2 & 50.4 & 61.9 & 49.0 & 38.7 & 65.3 & 52.0 & 13.75 & 7.50 \\
\midrule
\rowcolor{bandgray} \multicolumn{12}{c}{\bf Long Video MLLMs} \\
Video-XL-2-8B & 36.0 & 37.3 & 41.0 & 24.8 & 41.3 & 34.8 & 40.0 & 20.0 & 30.0 & 12.00$^*$ & 9.00$^*$ \\
InternVideo2.5-8B$^*$ & 34.4 & 37.3 & 42.6 & 30.4 & 31.7 & 34.8 & 28.0 & 21.3 & 24.7 & 13.75 & 5.25 \\
VideoChat-Flash-7B$^*$ & 36.8 & 39.7 & 34.4 & 31.2 & 41.3 & 36.4 & 33.3 & 34.7 & 34.0 & 12.50 & 5.25 \\
\midrule
\rowcolor{bandgray} \multicolumn{12}{c}{\bf Agentic Video MLLMs} \\
ReMA (GPT-5 agent) & -- & -- & -- & -- & -- & -- & -- & -- & -- & 18.82$^*$ & \underline{16.75}$^*$ \\
SiLVR-gpt-oss-120b & 35.2 & 42.9 & 57.4 & 52.8 & 57.1 & 47.0 & 22.7 & 60.0 & 41.3 & 9.50 & 6.00 \\
Ego-R1-Qwen3.5-35B & 33.6 & 38.1 & 37.7 & 34.4 & 50.8 & 37.6 & 30.7 & 29.3 & 30.0 & 14.25 & 6.50 \\
AVP-Qwen3.5-35B & 29.6 & 32.5 & 27.9 & 28.0 & 33.3 & 30.2 & 32.0 & 25.3 & 28.7 & 14.50 & 4.25 \\
WorldMM-GPT-5$^\dagger$ & 62.4 & 64.3 & \bf 75.4 & 62.4 & \underline{71.4} & 65.6 & -- & -- & -- & -- & -- \\
WorldMM-Qwen3.5-35B$^*$ & 49.6 & 54.8 & 54.1 & 62.4 & 60.3 & 56.0 & 48.0 & \underline{66.7} & 57.3 & \underline{31.42} & 8.50 \\
MAGIC-Video-Qwen3.5-35B$^*$ & \underline{67.2} & \underline{65.9} & 67.2 & \underline{66.4} & \bf 74.6 & \underline{67.6} & \underline{50.7} & \bf 78.7 & \underline{64.7} & 30.92 & 10.08 \\
\rowcolor{ourrow}
\name-Qwen3.5-35B (ours) & \bf 70.4 & \bf 73.8 & \underline{73.8} & \bf 71.2 & \underline{71.4} & \bf 72.0 & \bf 64.0 & \bf 78.7 & \bf 71.3 & \bf 36.83 & \bf 17.58 \\
\bottomrule
\end{tabular}
\vspace{-3mm}
\end{table}

\noindent\textbf{Week-long multiple-choice answering.}
\name{} achieves 72.0\% accuracy on EgoLifeQA, improving over the strongest competing memory framework, MAGIC-Video, by 4.4 percentage points. The corresponding gain on Ego-R1-Bench is 6.6 points with the same controller and answer model. Improvements over MAGIC-Video extend to four of the five EgoLifeQA question families, with the largest gain in EventRecall (+7.9 points, Table~\ref{tab:main}). Full system comparisons are provided in Appendix Table~\ref{tab:main-full}.

\noindent\textbf{Open-ended answering across recordings.}
On Test@Week, \name{} reaches 36.83\%, exceeding the strongest competing result, WorldMM, by 5.41 points. On the gameplay Test@Day split, its lead over the strongest competitor, ReMA, is 0.83 points. The reported confidence intervals exclude zero for EgoLifeQA, Ego-R1-Bench, and Test@Week, but include zero for Test@Day (Appendix Table~\ref{tab:significance}). \name{} exceeds both memory-framework baselines on each split. These results extend the evaluation to free-form answers and a recording domain with recurring game entities, with a clearer advantage on the week-long benchmark.

\noindent\textbf{Answering and temporal grounding.}
MultiHop-EgoQA additionally tests whether a system identifies the moments supporting its answer (Table~\ref{tab:multihop}). \name{} leads the compared memory frameworks on every reported metric: its answer score increases from WorldMM's 2.74 to 3.11, while mIoU increases by 1.6 points. \name{} also exceeds every model the benchmark reports in IoU@0.3 and mIoU, including GeLM, which is fine-tuned on the benchmark's training set. The improvements in both measures motivate examining which evidence reaches the answer model. The answer model reading 60 frames of the whole clip without a memory scores higher still, showing the gap that remains to answering from the complete clip.

\begin{table}[t]
\caption{\textbf{\name{} leads the compared memory frameworks on MultiHop-EgoQA.} Whole-clip models are separate references; Qwen3.5-35B reads 60 frames without memory. \emph{Full}: complete video for memory construction, with evidence retrieved at question time. IoU@0.3 averages all questions; mIoP, mIoG, and mIoU average those predicting intervals. \emph{Sim.}: sentence similarity. \emph{Score}: 1--10 grading by gpt-oss-120b, averaged over three runs for memory frameworks. $^\ddagger$: grounding and Sim.\ from \citet{chen2024multihop}, Score from released models using the same judge (Appendix~\ref{app:protocol}). \textbf{Bold}/\underline{underline}: best/second-best within each block.}
\label{tab:multihop}
\centering
\small
\setlength{\tabcolsep}{2.5pt}
\begin{tabular}{@{}lc@{\hspace{6pt}}cccc@{\hspace{6pt}}cc@{}}
\toprule
& & \multicolumn{4}{c}{\bf Temporal grounding} & \multicolumn{2}{c}{\bf Answering} \\
\cmidrule(lr){3-6} \cmidrule(lr){7-8}
Model & \# Frames & mIoP $\uparrow$ & mIoG $\uparrow$ & IoU@0.3 $\uparrow$ & mIoU $\uparrow$ & Sim. $\uparrow$ & Score $\uparrow$ \\
\midrule
Human$^\ddagger$ & Full & 71.8 & 81.0 & 87.0 & 61.8 & 74.3 & -- \\
\midrule
\rowcolor{bandgray} \multicolumn{8}{c}{\bf Whole-clip MLLMs} \\
GPT-4o$^\ddagger$ & 60 & 18.9 & 24.4 & 12.0 & 12.2 & \underline{73.7} & -- \\
InternVL2-8B$^\ddagger$ & 30 & 11.8 & 24.0 & 6.3 & 6.6 & 71.9 & \underline{3.33} \\
LLaVA-NeXT-Video-7B$^\ddagger$ & 32 & -- & -- & -- & -- & 62.1 & 2.91 \\
TimeChat-7B$^\ddagger$ & 96 & 10.2 & 5.6 & 3.0 & 3.6 & 58.9 & 2.24 \\
VTimeLLM-7B$^\ddagger$ & 100 & 12.4 & 28.2 & 8.8 & 9.2 & 70.5 & 2.95 \\
LLaVA-NeXT-7B $\rightarrow$ Llama-3.1-8B$^\ddagger$ & 180 & 21.4 & 22.3 & 10.1 & 9.7 & 63.6 & 2.23 \\
GeLM-7B$^\ddagger$ & 180 & \underline{23.7} & \underline{41.0} & \underline{18.2} & \underline{16.7} & \bf 75.0 & 3.19 \\
Qwen3.5-35B & 60 & \bf 33.7 & \bf 43.3 & \bf 33.1 & \bf 24.5 & 70.4 & \bf 3.64 \\
\midrule
\rowcolor{bandgray} \multicolumn{8}{c}{\bf Agentic Video MLLMs} \\
WorldMM-Qwen3.5-35B & Full & \underline{30.6} & \underline{34.2} & \underline{21.5} & \underline{18.9} & \underline{62.2} & \underline{2.74} \\
MAGIC-Video-Qwen3.5-35B & Full & 29.6 & 30.7 & 19.0 & 17.6 & 61.3 & 2.58 \\
\rowcolor{ourrow} \name-Qwen3.5-35B (ours) & Full & \bf 32.0 & \bf 36.7 & \bf 25.9 & \bf 20.5 & \bf 65.2 & \bf 3.11 \\
\bottomrule
\end{tabular}
\vspace{-6mm}
\end{table}

\subsection{Retrieval of Supporting Evidence}
\label{sec:exp-evidence}

An \emph{evidence hit} on EgoLifeQA occurs when a received caption or described observation overlaps the annotated evidence window. Observations contribute their description-context windows, which may extend beyond visible spans (Appendix~\ref{app:protocol}). MultiHop-EgoQA \emph{complete evidence coverage} (\emph{All} in Table~\ref{tab:multihop-coverage}) requires overlap with every annotated interval. These measures evaluate received context, while temporal grounding evaluates intervals predicted in answers. \emph{Retrieved duration} (\emph{Clip}) is the union of received intervals divided by duration, accounting for differences in the amount of retrieved context.

\begin{figure}[t]
  \centering
  \subfloat[][Evidence coverage \label{fig:analysis-coverage}]{%
    \includegraphics[width=0.32\linewidth]
      {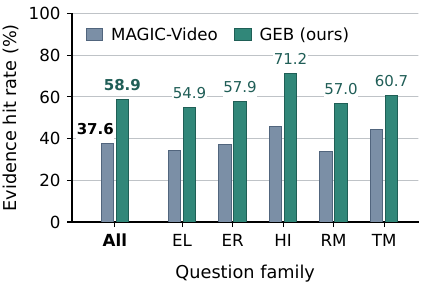}%
  }\hspace{0.06\linewidth}
  \subfloat[][Complete evidence \label{fig:analysis-multihop}]{%
    \includegraphics[width=0.32\linewidth]
      {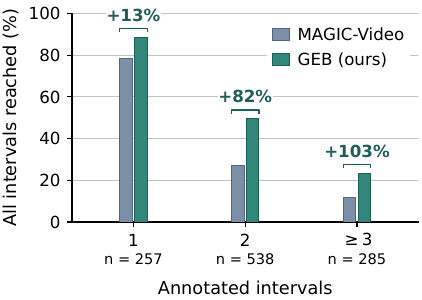}%
  }
  \caption{\textbf{\name{} improves access to annotated evidence over MAGIC-Video, with larger relative gains when evidence spans multiple intervals.}
    \textbf{(a)} Percentage of the 484 annotated EgoLifeQA questions whose evidence window overlaps a received unit, overall and by question family (abbreviations in Table~\ref{tab:main}).
    \textbf{(b)} Percentage of MultiHop-EgoQA questions for which received units overlap every annotated interval, grouped by interval count. Brackets show relative gains over MAGIC-Video, and $n$ gives the number of questions in each group.}
  \label{fig:analysis}
  \vspace{-4mm}
\end{figure}

\noindent\textbf{Access to relevant moments.}
Figure~\ref{fig:analysis}a shows that \name{} increases EgoLifeQA evidence hits from 37.6\% to 58.9\% compared with MAGIC-Video. The improvement holds across all question families. Biographies make an observation's source interval available alongside episodic evidence, providing additional routes to the moments a question concerns. 

\noindent\textbf{Access to distributed evidence.}
On MultiHop-EgoQA, complete evidence coverage increases from 35.4\% to 52.1\% over MAGIC-Video, a relative gain of 47\% (Table~\ref{tab:multihop-coverage}). Figure~\ref{fig:analysis}b and the \emph{All, $\geq$3} column show larger relative gains for questions requiring several intervals than for those requiring one. These gains support linking observations across events to recover distributed evidence.

\begin{table}[t]
\caption{\textbf{\name{} exceeds MAGIC-Video in complete coverage and is comparable to WorldMM with lower retrieved duration.} \emph{All}: fraction of questions with overlap for every annotated interval; \emph{All, $\geq$3}: the same over the 285 questions with three or more intervals. \emph{Clip}: union of received intervals divided by clip duration. Score and mIoU follow Table~\ref{tab:multihop}; Score gives mean$\pm$std across runs. $\Delta$ All: row minus \name{}, with a bootstrap $95\%$ confidence interval over questions. The lower block removes biography text or association.}
\label{tab:multihop-coverage}
\centering
\small
\setlength{\tabcolsep}{2pt}
\begin{tabular}{@{}lcc@{\hspace{4pt}}c@{\hspace{4pt}}cc@{\hspace{4pt}}c@{}}
\toprule
& \multicolumn{2}{c}{\bf Evidence reached} & {\bf Read} & \multicolumn{2}{c}{\bf Outcome} & {\bf vs.\ ours} \\
\cmidrule(lr){2-3} \cmidrule(lr){4-4} \cmidrule(lr){5-6} \cmidrule(lr){7-7}
Memory & All $\uparrow$ & All, $\geq$3 $\uparrow$ & Clip $\downarrow$ & Score $\uparrow$ & mIoU $\uparrow$ & $\Delta$ All [$95\%$ CI] \\
\midrule
MAGIC-Video-Qwen3.5-35B & 0.354 & 0.116 & 0.230 & 2.58$\pm$0.07 & 17.6 & $-0.167\;[-0.190,\,-0.145]$ \\
\quad 6 units/round & 0.449 & 0.182 & 0.319 & 2.75$\pm$0.05 & 18.5 & $-0.071\;[-0.096,\,-0.048]$ \\
WorldMM-Qwen3.5-35B & 0.526 & 0.244 & 0.399 & 2.74$\pm$0.06 & 18.9 & $+0.005\;[-0.019,\,+0.030]$ \\
\rowcolor{ourrow} \name-Qwen3.5-35B (ours) & 0.521 & 0.235 & 0.343 & 3.11$\pm$0.03 & 20.5 &  \\
\midrule
\multicolumn{7}{@{}l}{\emph{Ours with one decision removed}} \\
\quad w/o biography text (index only) & 0.468 & 0.164 & 0.279 & 3.06$\pm$0.06 & 19.8 & $-0.053\;[-0.070,\,-0.034]$ \\
\quad w/o association & 0.484 & 0.194 & 0.309 & 3.11$\pm$0.04 & 19.9 & $-0.037\;[-0.055,\,-0.019]$ \\
\bottomrule
\end{tabular}
\vspace{-3mm}
\end{table}

\noindent\textbf{Accounting for retrieved duration.}
To assess whether the gain follows simply from retrieving more of the recording, we increase MAGIC-Video's MultiHop-EgoQA allowance from three to six units per round. Complete evidence coverage remains below \name{} at a similar retrieved duration (Table~\ref{tab:multihop-coverage}). WorldMM achieves comparable complete coverage to \name{}, but its retrieved units span more of the clip and its answer score is lower. Thus, complete coverage alone does not explain answer quality. 

\subsection{Ablation Studies}
\label{sec:exp-ablation}

Table~\ref{tab:ablation} isolates memory organization, episodic connections, and reading inputs on EgoLifeQA with the controller and answer model fixed.

\begin{table}[t]
\caption{\textbf{\name{} benefits from physical-instance organization, contextual connections, and biography reading.} EgoLifeQA accuracy (\%) on 500 questions with fixed controller and answer model. Left block: the index. Right block: the reading. $\Delta$: variant minus full \name{}, in percentage points. Variant definitions: Section~\ref{sec:exp-ablation}; additional edge ablations: Table~\ref{tab:ablation-more}.}
\label{tab:ablation}
\centering
\small
\setlength{\tabcolsep}{1.6pt}
\begin{tabular}{@{}lcc@{\hspace{8pt}}lcc@{}}
\toprule
Memory & Acc. $\uparrow$ & $\Delta$ & Memory & Acc. $\uparrow$ & $\Delta$ \\
\midrule
\rowcolor{ourrow}
\name{} (full) & 72.0 & -- & & & \\
\addlinespace[3pt]
\multicolumn{3}{@{}l}{\emph{The index}} & \multicolumn{3}{@{}l}{\emph{The reading}} \\
\quad w/o association & 68.6 & $-3.4$ & \quad w/o biography text (index only) & 68.0 & $-4.0$ \\
\quad identity keyed by name & 69.2 & $-2.8$ & \quad w/o biography text for the answer model & 70.8 & $-1.2$ \\
\quad descriptions appended to captions & 68.2 & $-3.8$ & \quad w/o identity notes & 68.2 & $-3.8$ \\
\quad w/o observation$\to$timeline edges & 68.2 & $-3.8$ & \quad w/o unsearched-observation line & 70.2 & $-1.8$ \\
 & & & \quad w/o caption text & 61.8 & $-10.2$ \\
 & & & \quad w/o visual frames & 70.4 & $-1.6$ \\
\bottomrule
\end{tabular}
\vspace{-4mm}
\end{table}

\noindent\textbf{Physical identity beyond additional descriptions.}
Without association, each observation forms its own entity; name-keyed identity groups observations by described name. Both retain all descriptions under the full model's retrieval-unit and frame limits and deliver comparable context, but reduce accuracy by 3.4 and 2.8 points. Appending descriptions to captions removes observation and entity structure and costs 3.8 points at a matched answering-context budget. These controls support physical-instance organization beyond additional descriptions.

\noindent\textbf{Connections to identity and event context.}
Removing \textsc{same-instance} edges while retaining entity assignments costs 3.0 points. Disconnecting observations from episodes and source clips costs 3.8 points, and removing both edge families costs 5.0 (Appendix Table~\ref{tab:ablation-more}). These results support both following identity across events and recovering the context of each observation.

\noindent\textbf{Biographies for answering and further search.}
Withholding biography text from both models while retaining the indexed memory costs 4.0 points. Withholding it only from the answer model while keeping the controller's searches fixed reduces accuracy by 1.2 points. Removing references to observations not yet selected costs 1.8 points. The text-removal controls establish benefits beyond graph retrieval, while the reference removal supports using biographies to guide further search.

\noindent\textbf{Consistency across benchmarks.}
On MM-Lifelong, appending descriptions to captions costs 3.25 points on Test@Week and 5.08 on Test@Day; withholding biography text with the index retained costs 6.75 and 2.33, respectively. Neither variant recovers full performance on either split (Appendix Table~\ref{tab:mmlifelong-ablation}). On MultiHop-EgoQA, both association and biography-text removal reduce complete evidence coverage, with confidence intervals excluding zero, but have smaller effects on answer scores (Table~\ref{tab:multihop-coverage}). Appendix~\ref{app:rendered} illustrates how biographies and context resolve concrete questions.

\vspace{-3mm}
\section{Conclusion}
Long-video memory must preserve not only what happened, but also which people and objects connect events across time. We introduced Grounded Entity Biographies, a framework that links visually grounded observations of the same physical instance while retaining each observation’s context and source evidence. These biographies make persistent entities bridges across episodes, allowing retrieval to follow an entity's history and use previously recorded appearances to guide further search. Evaluations across four benchmarks demonstrate improvements in multiple-choice and open-ended question answering, including week-long recordings. Evidence analysis shows better access to relevant moments, while ablations support the contributions of grounded identity association and biography reading beyond additional descriptions alone. These findings highlight the value of organizing video memory around enduring subjects alongside the events in which they participate.

\ifarxiv\else
\clearpage
\subsection*{AI use statement}
Large language models were used to assist with editing and polishing the writing of this paper. All scientific content, experimental results, and claims were reviewed and verified by the authors, who take responsibility for the final content of the paper.

\subsection*{Ethics statement}
All recordings and questions used in this work come from publicly released benchmarks, and no new data were collected.

\subsection*{Reproducibility statement}
Section~\ref{sec:method} and Appendix~\ref{app:implementation} give every component, threshold and retrieval cap, and Appendix~\ref{app:protocol} the evaluation protocol for each benchmark and for the baselines run by us. The code and the artifacts behind the reported numbers will be released.
\fi

\bibliography{references}
\bibliographystyle{iclr2027_conference}

\clearpage
\appendix
\section*{Appendix}
\section{Implementation and Hyperparameters}
\label{app:implementation}

\paragraph{Video units and tracks.}
The EgoLife week is processed as 6,266 thirty-second clips of $1408\times1408$ video, sampled at 10 fps. YOLOE-26x-seg in its prompt-free mode detects objects on every sampled frame \citep{wang2025yoloe}, BoT-SORT links the detections into tracks within a clip \citep{aharon2022botsort,tang2023egotracks}, and duplicate tracks of one object are merged. Following VideoAgent \citep{fan2024videoagentmem}, tracks that the tracker split are regrouped by CLIP and DINOv2 appearance similarity \citep{radford2021clip,oquab2024dinov2}, and tracks seen in the same frame are never grouped. A group visible for at least two seconds becomes an observation and keeps representative crops, its time span, and its source episode; whole-frame and watermark boxes are discarded.

\paragraph{Description.}
Each observation is described by one call to a locally served vision-language model (\texttt{Qwen3.5-35B}). The call sees four crops at 448 px, scene frames at 768 px sampled at 1 fps over the observation's span (3 to 30) with the subject boxed, and the recording's own caption and transcript within $\pm40$ s. It returns the object's name, a one-paragraph summary that is used as the retrieval text, an event description, whether the object is a person, its location, and its stable attributes.

\paragraph{Association.}
\label{app:association}
Each observation is embedded with \texttt{Qwen3-VL-Embedding-8B} from its crops, name and summary. Every entity with a reference observation at or above $\theta_{\mathrm{match}}$ is a candidate, and acceptance follows Equation~\ref{eq:association} over the reference set of each entity, its 10 most recently assigned observations. A box that stays in place across the cut between two consecutive clips (the mutual best match with IoU at least 0.6) also counts as the match anchor, and the consistency floor and the separation test still apply. Two observations of one clip are visually separate ($D_{ij}=1$) when they share at least three frames and their boxes overlap with IoU $<0.5$ in at least two thirds of them. The two thresholds are set per recording by visual inspection of sampled merges: $\theta_{\mathrm{cons}},\theta_{\mathrm{match}}=0.60,0.75$ on EgoLife and $0.75,0.85$ on the gameplay stream, whose repeated assets make objects of one kind score alike.

\paragraph{Naming.}
One vision call per clip assigns the people on screen to the recording's cast jointly, and a text model (\texttt{gpt-oss-120b}) distills each person's clothing for the day from those assignments. An entity matched to one unambiguous profile takes its name and carries it to its observations on the other days, and an entity that two days name differently carries no name across days. The name labels the entity, and the description text is left as written. This stage runs on the EgoLife recording, whose seven participants are named; the Test@Day livestream has no cast, so its entities carry only the names written in their descriptions.

\paragraph{Graph and retrieval.}
Observation texts are embedded with the caption encoder (\texttt{Qwen3-Embedding-4B}, 2560-d) and indexed in the same lexical index as the captions. Edge weights are episode context 1.0, visual provenance 0.8, and \textsc{same-instance} 0.5. On EgoLife, the episodic memory uses 30-second, 3-minute, 10-minute, and 1-hour scales. Personalized PageRank uses a damping factor of 0.85, and each returned node is ranked by the product of its PageRank score and its query--text cosine similarity. An entity node exists for every entity with two or more observations. It routes relevance between the observations and is not scored as a unit itself. Each retrieved observation is one unit of the round's budget; the retrieved observations of one entity are rendered together as its biography excerpt, the \emph{Retrieved entity} block of Figure~\ref{fig:qualitative}. Retrieval on every benchmark allows at most five rounds and 64 frames. Week-scale retrieval uses 16 units per round, and at most six of the units may be observations, at most three of them from one entity and with no limit on the number of entities. MultiHop-EgoQA uses three units per round, of which at most one may be an observation. On every benchmark the controller and the answer model are \texttt{Qwen3.5-35B}, called either on a locally served vLLM instance, its context window extended to 1M tokens with YaRN, or through OpenRouter's \texttt{qwen3.5-flash} endpoint, the hosted form of the same checkpoint. \name{} and MAGIC-Video share one episodic memory: the 30-second captions and the topic and event summaries derived from them, injected after each search. The summaries are extracted once from the captions and do not depend on the entities. For a question asked at time $t$, the memory is re-indexed to the nodes before $t$.

\section{Memory Scale and Retrieval Cost}
\label{app:scale}

Table~\ref{tab:scale} summarizes the memory built for the EgoLife week, and Table~\ref{tab:graph-stats} counts the nodes and edges of the three memory frameworks on the graph a question at the end of the week retrieves over. At the selected operating point, association groups 257,974 observations ($83.7\%$ of all) into 27,446 entities of two or more observations, 15,365 of which span more than one day. Beside the shared episodic memory, the memory graph holds one node per tracked observation and one per entity with two or more observations, so its size follows the number of tracked observations, about 49 per 30-second clip. The memory is written once, offline, and serves every later question.

\begin{table}[!htb]
\caption{Scale of the memory built for the EgoLife week. Entities are counted once each, including those carried across several days; the seven named people are one entity each.}
\label{tab:scale}
\centering
\small
\begin{tabular}{@{}lr@{}}
\toprule
Quantity & Count \\
\midrule
Thirty-second video clips & 6,266 \\
Approximate video duration & 52 hours \\
Tracked object observations & 308,244 \\
Entities after association & 77,716 \\
\quad of which join two or more observations & 27,446 \\
\quad of which span more than one day & 15,365 \\
Observations inside entities of two or more observations & 257,974 \\
Largest entity, a named person / an unnamed object & 5,264 / 517 \\
\bottomrule
\end{tabular}
\end{table}

\paragraph{Name collisions in the corpus.}
The frozen memory lets us count how often a described name fails to identify an instance. Two observations of one clip are proven to be different objects by the same-frame separation test of Appendix~\ref{app:implementation}, which uses geometry alone and does not depend on the association. In the EgoLife week, 89,888 such pairs carry the same described name, and they occur in 5,804 of the 6,266 clips (92.6\%). In the other direction, under our association, 77.2\% of the 22,401 object entities seen more than once are described under two or more distinct names (an air conditioner is also an \emph{air conditioning unit} and a \emph{wall-mounted air conditioner}), so a memory keyed by the described name both merges different instances and splits one instance.

\begin{table}[!htb]
\caption{\textbf{Memory statistics for the EgoLife A1 week}, counted on the memory a question asked at the end of the week retrieves over. WorldMM is not one graph: it keeps a HippoRAG graph per caption granularity (passage and extracted-entity vertices, weighted relation edges) beside separate semantic-triple and visual-clip indices. \name{} and MAGIC-Video share the episodic memory (episode, clip and temporal edges). \name{} adds one node per tracked observation and one per entity with two or more observations. MAGIC-Video adds its text-derived entity and triple layer.}
\label{tab:graph-stats}
\centering
\small
\setlength{\tabcolsep}{4pt}
\begin{tabular}{@{}lrrr@{}}
\toprule
& \shortstack[c]{WorldMM-\\Qwen3.5-35B} & \shortstack[c]{MAGIC-Video-\\Qwen3.5-35B} & \shortstack[c]{\name-Qwen3.5-35B\\(ours)} \\
\midrule
\rowcolor{bandgray} \multicolumn{4}{l}{\bf Nodes} \\
Episode nodes (captions at 4 granularities) & 7,624 & 7,625 & 7,625 \\
Visual clip nodes & 6,223$^\dagger$ & 6,223 & 6,223 \\
Named-entity nodes extracted from text & 55,468 & 2,669 & -- \\
Semantic triple nodes & 3,821$^\dagger$ & 3,821 & -- \\
Observation nodes & -- & -- & 308,244 \\
Entity nodes & -- & -- & 27,446$^\ddagger$ \\
Total graph nodes & 63,092 & 20,338 & 349,538 \\
\midrule
\rowcolor{bandgray} \multicolumn{4}{l}{\bf Edges} \\
Temporal adjacency (episode$\to$episode) & -- & 15,182 & 15,182 \\
Granularity containment (coarse$\to$fine episode) & -- & 13,560 & 13,560 \\
Episode$\leftrightarrow$clip & -- & 12,446 & 12,446 \\
Entity or observation$\to$episode (episode context) & -- & 18,619 & 306,233 \\
Entity or observation$\to$clip (visual provenance) & -- & 18,619 & 306,233 \\
Entity$\to$triple (has property) & -- & 4,020 & -- \\
Entity$\leftrightarrow$observation (\textsc{same-instance}) & -- & -- & 257,974 \\
HippoRAG passage--entity and relation edges & 565,096 & -- & -- \\
Total graph edges & 565,096 & 82,446 & 911,628 \\
\bottomrule
\end{tabular}

\smallskip
{\footnotesize $^\dagger$ Held in a separate index, not in the HippoRAG graphs; not counted in the WorldMM totals. $^\ddagger$ One node per entity with two or more observations. A single-observation entity is reached through its observation node.}
\end{table}

\paragraph{Retrieval cost.}
\label{app:latency}
We measure the query-time cost of the memory of \name{} on the 50 Ego-R1-Bench questions. The searches issued in an evaluation in the setting of Section~\ref{sec:exp-setup} are run again on one A100, without the controller and the answer model, against the memory of the whole week, which is at least as large as the memory any question sees, so the times below are upper bounds. Each search traverses the memory graph of Table~\ref{tab:graph-stats} with no language-model call, and its time covers retrieval and the formatting of the returned context. A search takes 5.6 s on average (median 5.1 s, 90th percentile 8.6 s), and a question, with 2.42 searches, spends 13.6 s in retrieval.

\section{Evaluation Protocol}
\label{app:protocol}

\paragraph{EgoLifeQA and Ego-R1-Bench.}
Both benchmarks use the protocol under which the published numbers in Table~\ref{tab:main} were obtained \citep{li2026magic}: participant A1, with the recording up to the query time visible (Section~\ref{sec:exp-setup}). Two EgoLifeQA questions (71 and 73) cannot be answered from their options, one with duplicated options and one whose gold option is identical to another, and count as wrong for every system we run. Multiple-choice answers are scored by exact match.

\paragraph{EgoLifeQA evidence coverage.}
Figure~\ref{fig:analysis}a evaluates the 484 questions with an annotated evidence window. We count an evidence hit when that window overlaps a 30-second caption or a described observation received by the answer model. An observation contributes the contextual window used to generate its description (Appendix~\ref{app:implementation}). Coarser episode summaries are not counted in this measure. Coverage measures access to temporally relevant context, not the correctness of the observation's identity assignment.

\paragraph{MultiHop-EgoQA.}
The benchmark releases 1,080 questions over 360 three-minute segments of Ego4D videos (854$\times$480, 30 fps, no audio track). MAGIC-Video, WorldMM and \name{} build their memory on 10-second units from the same captions and visual units. MAGIC-Video adds its name-keyed entities and triples, and WorldMM its separate episodic, semantic and visual stores. The multi-granularity aggregation and the chains summarize hours and have no content on a three-minute clip, so the shared episodic memory has a single granularity there. Our observations are tracked and described on 10-second clips from their own frames only, and associated across the clip's 18 segments by the same visual association as at week scale. A 10-second observation is one unit of the timeline it links to. \name{} retrieves three units per round, of which at most one may be an observation, for at most five rounds under the frame cap in Appendix~\ref{app:implementation}. MAGIC-Video also retrieves three units per round, whereas WorldMM retrieves up to three units from each of its episodic, semantic, and visual stores. The additional MAGIC-Video comparison increases its allowance to six units per round while retaining the controller, answer model, and scoring protocol. Because these allowances need not produce equal amounts of context, Table~\ref{tab:multihop-coverage} also reports the fraction of the clip covered by received units. A clip is 18 episodes, so 16 units would return most of it in one round. The controller and the answer model are those of Section~\ref{sec:exp-setup}, and frames enter the answering context only inside a retrieved visual unit.

\paragraph{MultiHop-EgoQA scoring.}
Answers follow the benchmark's open-ended prompt and are parsed and scored with its released code. mIoP, mIoG and mIoU compare the predicted evidence intervals with the annotation, averaged over the questions that predicted an interval, and IoU@0.3 is computed over all questions. Both come from the released script run unchanged on our predictions. Sentence similarity compares the answer with the reference (all-MiniLM-L6-v2). The benchmark's 1--10 grading prompt scores the 724 questions with a worded reference and runs on \texttt{gpt-oss-120b} rather than the answer model. The grounding metrics and the sentence similarity involve no judge, so the rows Table~\ref{tab:multihop} copies from the benchmark paper are on the same scale as ours in those columns. Among those rows, Human is scored on 10\% of the questions, GeLM is fine-tuned on the benchmark's training set, and the caption pipeline reads all 180 captions of a clip. Published grading scores use GPT-4o and are not reported. Instead we ran the released inference code and weights of every open-source system (InternVL2-8B, LLaVA-NeXT-Video-7B, TimeChat-7B, VTimeLLM-7B, the LLaVA-NeXT caption pipeline with Llama-3.1-8B, and GeLM-7B on its released features) and scored their answers with the same gpt-oss-120b judge. Their grounding metrics and sentence similarity match the reported values within one point for every system except the caption pipeline, whose mIoP and mIoG are two to three points below the reported values. GPT-4o itself is not re-run, so every comparison of grading scores in the paper is on one judge. Evidence coverage is read off the answering context: a caption unit or a described observation covers an evidence interval when their windows overlap. The confidence intervals of Table~\ref{tab:multihop-coverage} are computed over all 1,080 questions, each question's value being its mean over the three runs, and the judge score is averaged over the 724 questions with a worded reference.

\paragraph{MM-Lifelong.}
Test@Week is the same EgoLife recording as our main experiments, with 200 human-written questions and no query timestamp, so the whole week is observable and the memory built for EgoLifeQA is used unchanged. Test@Day is a 23.6-hour gameplay livestream with 200 questions. The three memory frameworks share its 30-second captions, which name the game and copy on-screen names because the questions refer to the game's entities by name. Static interface overlays (health bars, icons, the streamer's name tag) are not physical objects and are excluded from the memory, and the association thresholds are set by the inspection of Appendix~\ref{app:implementation}. Answers are free text, and the benchmark's GPT-5 judge scores each 0--5, mapped to 1, 0.5, or 0. The Qwen3.5-35B row of Table~\ref{tab:main} is the answer model alone. It reads 1536 frames sampled uniformly over the whole recording, the frame count of the published Qwen3-VL rows, sent as one video through the model's own video processor with each frame kept at 512 pixels on its longer side, without transcript or thinking and under the same judge. Table~\ref{tab:mmlifelong-nomem} varies the input of the answer model alone, and Table~\ref{tab:main} reports the strongest of these settings.

\begin{table}[!htb]
\caption{\textbf{Qwen3.5-35B alone on MM-Lifelong} The answer model of every memory framework we run reads the recording directly, its inputs sampled uniformly over the whole recording, 200 questions per cell. The first two rows send 1536 frames as one video through the model's own video processor, at 512 pixels on the longer side and at the processor's default budget of 160$\times$96 per frame. The third row sends 256 frames as separate images. The 64-frame rows use the protocol of \citet{li2026magic} for general MLLMs. Captions are the memory frameworks' 30-second captions at the sampled positions, and the frame-only rows carry no transcript.}
\label{tab:mmlifelong-nomem}
\centering
\small
\setlength{\tabcolsep}{6pt}
\begin{tabular}{@{}lcc@{}}
\toprule
Input & Test@Week $\uparrow$ & Test@Day $\uparrow$ \\
\midrule
1536 frames, longer side 512 pixels (the Qwen3.5-35B row of Table~\ref{tab:main}) & 13.75 & 7.50 \\
1536 frames at the processor's default budget, 160$\times$96 per frame & 9.25 & 6.50 \\
256 frames as images at 151,200 pixels each, greedy & 12.50 & 7.25 \\
64 frames, longest edge 512 & 11.50 & 7.00 \\
64 frames and their 64 captions & 9.00 & 4.75 \\
512 captions, no frames & 9.00 & 3.25 \\
\bottomrule
\end{tabular}
\end{table}

\paragraph{Baselines run by us.}
Unmarked entries in Tables~\ref{tab:main}, \ref{tab:main-full}, and~\ref{tab:mmlifelong-full} are our runs. These runs access only recordings preceding the query time on EgoLifeQA and Ego-R1-Bench, and the complete recording on MM-Lifelong.
On EgoLifeQA and Ego-R1-Bench the Qwen3.5-35B row reads 256 frames sampled up to the query time together with their captions. The long-video models (LongVA-7B \citep{zhang2024longva}, InternVideo2.5-8B \citep{wang2025internvideo25}, VideoLLaMA3-7B \citep{zhang2025videollama3}, Molmo2-8B \citep{clark2026molmo2}, Video-XL-2-8B \citep{qin2025videoxl2} and VideoChat-Flash-7B \citep{li2025videochatflash}) run from their public checkpoints with their own inference code and read 256 frames sampled uniformly at 1\,fps under one open-ended prompt. VideoChat-Flash-7B receives the frames as a 1\,fps video. The agentic rows run every text role on gpt-oss-120b and every frame-reading role on Qwen3.5-35B. SiLVR \citep{zhang2026silvr} hands every 30-second caption to one reasoning model, thinned to a character budget that fits the model's 128k-token context, and asks for the answer. AVP \citep{wang2025avp} is a plan-observe-reflect agent (three rounds, 512 frames per step at 384 pixels, a 1,024-frame budget) whose synthesizer returns a short answer. Ego-R1 \citep{tian2025egor1} runs its released prompt, tool schemas and twelve-turn loop, with the fine-tuned agent replaced by gpt-oss-120b and the video and frame tools reading our 1\,fps frames with Qwen3.5-35B. It uses its released hierarchical caption database for Test@Week and, for Test@Day, a database of the same three levels written by gpt-oss-120b from our 30-second captions. EgoButler \citep{yang2025egolife}, in Table~\ref{tab:main-full} only, runs EgoLife's released EgoRAG code unchanged on the same 30-second captions, every call on gpt-oss-120b. MM-Lifelong answers are scored by the benchmark's GPT-5 judge.

\section{Additional Results}
\label{app:results}

\paragraph{Ablation settings and context budgets.}
The EgoLifeQA variants without association and with identity keyed by name retain every observation description as a searchable unit under the full model's retrieval-unit and frame limits (Appendix~\ref{app:implementation}). For descriptions appended to captions, the observation and entity nodes are removed, and the retrieval allowance is adjusted to approximately match the full model's mean answering-context token and frame counts. Matching therefore concerns the delivered context, rather than the number of retrieved units. Withholding biography text retains the indexed observations and graph connections. The index-only variant withholds that text from both models, while the answer-model-only variant keeps the controller's searches fixed and removes it only from the final answering context.

\paragraph{Answering context.}
Table~\ref{tab:context} reports the answering context \name{} hands the answer model on EgoLifeQA with and without the visual frames, under the limits in Appendix~\ref{app:implementation}. Without frames, the answer model retains the textual biographies and captions and reads 8.2k tokens per question, with accuracy reported in the \emph{w/o visual frames} row of Table~\ref{tab:ablation}. This removal changes the input modality as well as context size. The separate \emph{w/o identity notes} variant removes the notes distinguishing visually separate same-named entities from pairs whose identity remains unresolved (Appendix~\ref{app:rendered}).

\begin{table}[!htb]
\caption{\textbf{Answering-context size with and without visual frames.} \emph{Tok} and \emph{Fr}: mean answering-context tokens and frame references per question on EgoLifeQA. The second row is the \emph{w/o visual frames} row of Table~\ref{tab:ablation}; controller, prompts, caps and video are the same.}
\label{tab:context}
\centering
\small
\setlength{\tabcolsep}{8pt}
\begin{tabular}{@{}lcc@{}}
\toprule
Memory & Tok & Fr \\
\midrule
\rowcolor{ourrow}
\name-Qwen3.5-35B (ours) & 130.0k & 62.9 \\
\quad w/o visual frames & 8.2k & 0.0 \\
\bottomrule
\end{tabular}
\end{table}

\paragraph{MM-Lifelong.}
Table~\ref{tab:mmlifelong-ablation} tests on both MM-Lifelong splits whether the gain comes from indexing the observations by identity or from their words: \emph{descriptions appended to captions} keeps the words without the index, and \emph{w/o biography text} keeps the index without its words. Both rows are defined as in Table~\ref{tab:ablation}, and each is the mean over three runs under the benchmark's judge. Table~\ref{tab:mmlifelong-full} lists every published row beside every system we ran.

\begin{table}[!htb]
\caption{\textbf{Withholding the biography text or appending the descriptions to the captions lowers the score on both splits.} MM-Lifelong accuracy (\%) under the benchmark's GPT-5 judge, mean $\pm$ std over three runs. The ablation rows are defined as in Table~\ref{tab:ablation}; the two memory frameworks run under our protocol repeat the means of Table~\ref{tab:main} with their std. $\Delta$: difference to the full memory.}
\label{tab:mmlifelong-ablation}
\centering
\small
\setlength{\tabcolsep}{5pt}
\begin{tabular}{@{}lcccc@{}}
\toprule
& \multicolumn{4}{c}{\bf MM-Lifelong $\uparrow$} \\
\cmidrule(lr){2-5}
Memory & Week & $\Delta$ & Day & $\Delta$ \\
\midrule
\rowcolor{ourrow} \name-Qwen3.5-35B (ours) & 36.83 $\pm$ 0.88 & -- & 17.58 $\pm$ 0.14 & -- \\
\quad w/o biography text (index only) & 30.08 $\pm$ 1.66 & $-6.75$ & 15.25 $\pm$ 1.15 & $-2.33$ \\
\quad descriptions appended to captions & 33.58 $\pm$ 0.80 & $-3.25$ & 12.50 $\pm$ 1.09 & $-5.08$ \\
\addlinespace[2pt]
MAGIC-Video-Qwen3.5-35B & 30.92 $\pm$ 0.80 & $-5.91$ & 10.08 $\pm$ 0.52 & $-7.50$ \\
WorldMM-Qwen3.5-35B & 31.42 $\pm$ 2.27 & $-5.41$ & 8.50 $\pm$ 2.63 & $-9.08$ \\
\bottomrule
\end{tabular}
\end{table}

\begin{table}[!htb]
\caption{MM-Lifelong answer accuracy (\%) with every published row of \citet{chen2026mmlifelong} (their Tables 4 and 15) and every system we ran; entries marked with $^*$ are taken from the original paper; MAGIC-Video, WorldMM and \name{} report the mean over three runs; the rows without $^*$ are run by us: the long-video models read 256 frames sampled uniformly at 1\,fps with their released scripts. \textbf{Bold} marks the best number within each group.}
\label{tab:mmlifelong-full}
\centering
\small
\setlength{\tabcolsep}{8pt}
\begin{tabular}{@{}lccc@{}}
\toprule
Model & \# Frames & Test@Week $\uparrow$ & Test@Day $\uparrow$ \\
\midrule
Human$^*$                 & Full & 95.6 & 99.2 \\
\midrule
\rowcolor{bandgray} \multicolumn{4}{c}{\bf General MLLMs} \\
GPT-5$^*$ & 50 & 15.00 & \bf 15.25 \\
Qwen3-VL-235B$^*$ & 1536 & \bf 15.63 & 12.44 \\
Qwen3-VL-30B$^*$ & 1536 & 11.07 & 11.48 \\
Qwen3.5-35B & 1536 & 13.75 & 7.50 \\
\midrule
\rowcolor{bandgray} \multicolumn{4}{c}{\bf Long Video MLLMs} \\
Video-XL-2-8B$^*$ & 2048 & 10.25 & 8.75 \\
Video-XL-2-8B$^*$ & 1024 & 12.00 & 9.00 \\
Eagle-2.5-8B$^*$ & 512 & 9.50 & 7.25 \\
Eagle-2.5-8B$^*$ & 32 & 7.00 & 8.25 \\
Nemotron-v2-12B$^*$ & 512 & 11.00 & 7.25 \\
Nemotron-v2-12B$^*$ & 128 & 8.50 & 7.00 \\
LongVA-7B & 256 & 5.00 & 5.00 \\
InternVideo2.5-8B & 256 & \bf 13.75 & 5.25 \\
VideoLLaMA3-7B & 256 & 11.25 & 5.75 \\
Molmo2-8B & 256 & 9.50 & \bf 10.00 \\
VideoChat-Flash-7B & 256 & 12.50 & 5.25 \\
\midrule
\rowcolor{bandgray} \multicolumn{4}{c}{\bf Agentic Video MLLMs} \\
VideoMind-7B$^*$ & Full & 11.75 & 7.50 \\
LongVT-7B$^*$ & Full & 9.75 & 7.00 \\
DeepVideoDiscovery$^*$ & Full & 9.02 & 10.25 \\
ReMA (GPT-5 agent)$^*$ & Full & 18.82 & 16.75 \\
ReMA (Qwen3-VL-235B agent)$^*$ & Full & 15.98 & 13.33 \\
SiLVR-gpt-oss-120b & Full & 9.50 & 6.00 \\
Ego-R1-Qwen3.5-35B & Full & 14.25 & 6.50 \\
AVP-Qwen3.5-35B & Full & 14.50 & 4.25 \\
WorldMM-Qwen3.5-35B & Full & 31.42 & 8.50 \\
MAGIC-Video-Qwen3.5-35B & Full & 30.92 & 10.08 \\
\rowcolor{ourrow}
\name-Qwen3.5-35B (ours) & Full & \bf 36.83 & \bf 17.58 \\
\bottomrule
\end{tabular}
\end{table}

\paragraph{MultiHop-EgoQA.}
Table~\ref{tab:multihop-official} reports results on all six benchmark metrics for the memory frameworks, the six-unit volume control, and the two ablation variants in Table~\ref{tab:multihop-coverage}. Table~\ref{tab:multihop-category} breaks down the judge scores by question category. Table~\ref{tab:multihop-rounds} tracks complete evidence coverage after each search round, as measured from the controller's search logs. The coverage gap between \name{} and the variant without biography text widens from 2.5 percentage points after the first round to 5.1 after the fifth.

\begin{table}[!htb]
\caption{\textbf{Under the benchmark's released metrics, the three baseline rows and both variants lie below the full memory on mIoG, IoU@0.3, mIoU and Sim.} Columns as in Table~\ref{tab:multihop}, Score as the mean $\pm$ std over three runs. The upper block holds the memory frameworks of Table~\ref{tab:multihop} and the six-unit volume control of Section~\ref{sec:exp-evidence}. The lower block removes one of the two decisions the memory rests on, as in Table~\ref{tab:multihop-coverage}.}
\label{tab:multihop-official}
\centering
\small
\setlength{\tabcolsep}{4pt}
\begin{tabular}{@{}lcccccc@{}}
\toprule
& \multicolumn{4}{c}{\bf Temporal grounding} & \multicolumn{2}{c}{\bf Answering} \\
\cmidrule(lr){2-5} \cmidrule(lr){6-7}
Memory & mIoP $\uparrow$ & mIoG $\uparrow$ & IoU@0.3 $\uparrow$ & mIoU $\uparrow$ & Sim. $\uparrow$ & Score $\uparrow$ \\
\midrule
MAGIC-Video-Qwen3.5-35B & 29.6 & 30.7 & 19.0 & 17.6 & 61.3 & 2.58$\pm$0.07 \\
MAGIC-Video-Qwen3.5-35B, 6 units/round & 30.2 & 33.2 & 20.7 & 18.5 & 63.1 & 2.75$\pm$0.05 \\
WorldMM-Qwen3.5-35B & 30.6 & 34.2 & 21.5 & 18.9 & 62.2 & 2.74$\pm$0.06 \\
\rowcolor{ourrow} \name-Qwen3.5-35B (ours) & 32.0 & 36.7 & 25.9 & 20.5 & 65.2 & 3.11$\pm$0.03 \\
\midrule
\multicolumn{7}{@{}l}{\emph{Ours with one decision removed}} \\
\quad w/o biography text (index only) & 32.1 & 34.2 & 24.4 & 19.8 & 64.8 & 3.06$\pm$0.06 \\
\quad w/o association & 32.5 & 34.1 & 25.3 & 19.9 & 64.8 & 3.11$\pm$0.04 \\
\bottomrule
\end{tabular}
\end{table}

\begin{table}[!htb]
\caption{\textbf{MultiHop-EgoQA judge score by question category} (mean over three runs; $n$ = judged questions). Categories follow the benchmark's annotation; the smaller categories hold 34 to 73 questions.}
\label{tab:multihop-category}
\centering
\small
\setlength{\tabcolsep}{1pt}
\begin{tabular}{@{}llccccc@{}}
\toprule
& & \multicolumn{5}{c}{Score $\uparrow$} \\
\cmidrule(lr){3-7}
Category & $n$ & \shortstack[c]{MAGIC-Video-\\Qwen3.5-35B} & \shortstack[c]{MAGIC-Video-\\Qwen3.5-35B,\\6 units/round} & \shortstack[c]{WorldMM-\\Qwen3.5-35B} & \shortstack[c]{\name-Qwen3.5-35B\\(ours)} & \shortstack[c]{Qwen3.5-35B,\\all 60 frames} \\
\midrule
A: Repeated activities & 73 & 2.91 & 3.00 & 2.97 & 3.24 & 4.10 \\
B: Multiple actions & 241 & 2.53 & 2.73 & 2.75 & 2.91 & 3.22 \\
C: Multiple objects & 193 & 2.40 & 2.51 & 2.43 & 2.90 & 3.47 \\
D: Locations/people & 72 & 2.51 & 2.79 & 2.59 & 3.25 & 3.85 \\
E: Event composition & 111 & 2.32 & 2.46 & 2.57 & 2.90 & 3.34 \\
F: Event comparison & 34 & 4.23 & 4.61 & 4.68 & 5.90 & 7.09 \\
\bottomrule
\end{tabular}
\end{table}

\begin{table}[!htb]
\caption{\textbf{Evidence reached after each search round on MultiHop-EgoQA}: the fraction of questions whose every evidence interval had been retrieved after the controller's first $r$ searches (mean over three runs, read off the controller's search log as \emph{All} in Table~\ref{tab:multihop-coverage} is read off the answering context). A question that stops early keeps its final value.}
\label{tab:multihop-rounds}
\centering
\small
\setlength{\tabcolsep}{5pt}
\begin{tabular}{@{}lccccc@{}}
\toprule
& \multicolumn{5}{c}{All $\uparrow$ after $r$ searches} \\
\cmidrule(lr){2-6}
Memory & $r=1$ & $r=2$ & $r=3$ & $r=4$ & $r=5$ \\
\midrule
MAGIC-Video-Qwen3.5-35B & 0.171 & 0.251 & 0.296 & 0.331 & 0.354 \\
MAGIC-Video-Qwen3.5-35B, 6 units/round & 0.235 & 0.326 & 0.386 & 0.420 & 0.449 \\
WorldMM-Qwen3.5-35B & 0.294 & 0.363 & 0.436 & 0.489 & 0.526 \\
\rowcolor{ourrow} \name-Qwen3.5-35B (ours) & 0.287 & 0.387 & 0.455 & 0.497 & 0.519 \\
\midrule
\multicolumn{6}{@{}l}{\emph{Ours with one decision removed}} \\
\quad w/o biography text (index only) & 0.262 & 0.353 & 0.402 & 0.442 & 0.468 \\
\quad w/o association & 0.279 & 0.373 & 0.424 & 0.462 & 0.484 \\
\bottomrule
\end{tabular}
\end{table}

\clearpage
\paragraph{Statistical significance of the headline gains.}
Table~\ref{tab:significance} gives 95\% confidence intervals on the gain of \name{} over the strongest baseline for each overall benchmark score in Table~\ref{tab:main}, following the procedure of \citet{li2026magic}. Where the anchor is a published number or a mean under our protocol, the baseline's per-question scores are not paired with ours, so the interval is a one-sample bootstrap of \name{}'s own per-question scores (2,000 resamples) with the anchor subtracted as a constant; it reflects the sampling spread of \name{} alone and can be asymmetric. On Ego-R1-Bench both systems are evaluated over three seeds, and the interval adds MAGIC-Video's reported per-seed variance to ours in quadrature. The intervals on EgoLifeQA, Ego-R1-Bench and Test@Week exclude 0. On Test@Day the strongest baseline is ReMA's reported 16.75, and the interval on the 0.83-point gain includes 0; against MAGIC-Video and WorldMM on that split (10.08 and 8.50) the same bootstrap gives [+3.34, +11.84] and [+4.83, +13.50]. The remaining tables give the full layouts: Table~\ref{tab:ablation-more} the EgoLifeQA ablation rows that Section~\ref{sec:exp-ablation} does not discuss, and Table~\ref{tab:main-full} the full EgoLifeQA and Ego-R1-Bench comparison behind Table~\ref{tab:main}, with the frame budget and modality of every system and the published rows on the same question sets.

\begin{table}[!htb]
\caption{\textbf{95\% confidence intervals on the headline gains of \name{} over the strongest baseline of each column of Table~\ref{tab:main}.} EgoLifeQA and MM-Lifelong: one-sample bootstrap (2,000 resamples) of \name{}'s per-question scores, the anchor subtracted as a constant; on MM-Lifelong a question's score is its mean over our three runs. Where \name{} has three runs its cell is the mean $\pm$ standard deviation over them; on Ego-R1 the interval adds the anchor's reported per-seed variance to ours in quadrature. $^*$: the baseline's published number; without a marker, its mean over three runs under Section~\ref{sec:exp-setup}.}
\label{tab:significance}
\centering
\small
\setlength{\tabcolsep}{2.5pt}
\begin{tabular}{@{}llccc@{}}
\toprule
Benchmark ($n$) & Strongest baseline & \name{} & Gap & 95\% CI \\
\midrule
EgoLifeQA (500) & MAGIC-Video-Qwen3.5-35B$^*$ (67.6) & 72.0 & +4.4 & [+0.4, +8.4] \\
Ego-R1-Bench (50 $\times$ 3) & MAGIC-Video-Qwen3.5-35B$^*$ (64.7 $\pm$ 1.15) & 71.3 $\pm$ 3.1 & +6.6 & [+2.9, +10.3] \\
Test@Week (200 $\times$ 3) & WorldMM-Qwen3.5-35B (31.42) & 36.83 $\pm$ 0.88 & +5.41 & [+0.16, +10.92] \\
Test@Day (200 $\times$ 3) & ReMA (GPT-5 agent)$^*$ (16.75) & 17.58 $\pm$ 0.14 & +0.83 & [$-$3.42, +5.58] \\
\bottomrule
\end{tabular}
\end{table}

\begin{table}[!htb]
\caption{\textbf{Removing the relevance propagation along a biography costs three points, and removing both edge families five.} Further ablations on EgoLifeQA (accuracy, \%), under the setting of Table~\ref{tab:ablation}. \emph{w/o} \textsc{same-instance} \emph{edges} keeps the entity grouping and removes only the propagation of relevance between an entity's observations; the combined row removes both edge families of Section~\ref{sec:method-schema}.}
\label{tab:ablation-more}
\centering
\small
\setlength{\tabcolsep}{4pt}
\begin{tabular}{@{}lcc@{}}
\toprule
Memory & Acc. $\uparrow$ & $\Delta$ \\
\midrule
\rowcolor{ourrow}
\name{} (full) & 72.0 & -- \\
\quad w/o \textsc{same-instance} edges & 69.0 & $-3.0$ \\
\quad w/o \textsc{same-instance} and observation$\to$timeline edges & 67.0 & $-5.0$ \\
\bottomrule
\end{tabular}
\end{table}

\begin{table}[!htb]
\caption{\textbf{Full comparison on EgoLifeQA and Ego-R1-Bench} (accuracy, \%), with the frame budget and modality of every system. A mark on a model name gives the paper its numbers are taken from, each with its own answer model: $^*$ \citet{li2026magic}, $^\dagger$ \citet{yeo2025worldmm}, which ran the marked systems itself, $^\ddagger$ \citet{rege2026egagent}, $^\S$ \citet{yang2025egolife}. Unmarked rows are our runs, with frame budgets and input modalities shown in the table. Ego-R1-Bench results are averaged over three runs (Appendix~\ref{app:protocol}). Numbers reported on other question sets are omitted. \textbf{Bold}/\underline{underline}: best/second-best in each column. Families as in Table~\ref{tab:main}.}
\label{tab:main-full}
\centering
\small
\setlength{\tabcolsep}{1.5pt}
\renewcommand{\arraystretch}{1.0}
\begin{tabular}{@{}lcc@{\hspace{4pt}}cccccc@{\hspace{4pt}}ccc@{}}
\toprule
& & & \multicolumn{6}{c}{\bf EgoLifeQA $\uparrow$} & \multicolumn{3}{c}{\bf Ego-R1 $\uparrow$} \\
\cmidrule(lr){4-9} \cmidrule(lr){10-12}
Model & \# Frames & Modality
& EL & ER & HI & RM & TM & Avg.
& Manual & Gemini & Avg. \\
& & & \tiny 125 & \tiny 126 & \tiny 61 & \tiny 125 & \tiny 63 & \tiny 500
      & \tiny 25 & \tiny 25 & \tiny 50 \\
\midrule
\rowcolor{bandgray} \multicolumn{12}{c}{\bf General MLLMs} \\
\multirow{3}{*}{Qwen3.5-9B$^*$}
                        & 64   & V   & 32.8 & 30.2 & 45.9 & 28.8 & 20.6 & 31.2 & 24.0 & 42.7 & 33.3 \\
                        & 512  & T   & 25.6 & 28.6 & 47.5 & 35.2 & 41.3 & 33.4 & 33.3 & 46.7 & 40.0 \\
                        & 64   & V+T & 28.8 & 31.7 & 45.9 & 33.6 & 36.5 & 33.8 & 22.7 & 64.0 & 43.3 \\
\midrule
Qwen3.5-35B                 & 256  & V+T & 43.2 & 46.8 & 49.2 & 50.4 & 61.9 & 49.0 & 38.7 & 65.3 & 52.0 \\
Qwen3-VL-235B               & 256  & V+T & 44.0 & 44.4 & 50.8 & 44.8 & 50.8 & 46.0 & 40.0 & 62.7 & 51.3 \\
GPT-5.4 Mini$^*$            & 64   & V+T & 37.6 & 40.5 & 44.3 & 34.4 & 41.3 & 38.8 & 22.7 & 40.0 & 31.3 \\
Gemini 3.1 Flash Lite$^*$   & 64   & V+T & 44.0 & 41.3 & 42.6 & 47.2 & 38.1 & 43.2 & 29.3 & 66.7 & 48.0 \\
GPT-4.1$^\ddagger$          & --   & T   & 32.0 & 39.7 & 39.3 & 32.8 & 39.7 & 36.0 & --   & --   & --   \\
Gemini 2.5 Pro$^\ddagger$   & 3000 & V+T & 45.6 & 48.4 & 51.7 & 41.6 & 52.4 & 46.8 & --   & --   & --   \\
GPT-5$^\dagger$             & --   & --  & 47.2 & 42.1 & 47.5 & 53.6 & 55.6 & 48.6 & --   & --   & --   \\
\midrule
\rowcolor{bandgray} \multicolumn{12}{c}{\bf Long Video MLLMs} \\
VideoLLaMA3-7B$^*$          & 128  & V   & 32.8 & 35.7 & 37.7 & 27.2 & 33.3 & 32.8 & 34.7 & 36.0 & 35.3 \\
InternVideo2.5-8B$^*$       & 512  & V   & 34.4 & 37.3 & 42.6 & 30.4 & 31.7 & 34.8 & 28.0 & 21.3 & 24.7 \\
LongVA-7B$^*$               & 128  & V   & 33.6 & 38.1 & 36.1 & 39.2 & 28.6 & 35.8 & 28.0 & 18.7 & 23.3 \\
VideoChat-Flash-7B$^*$      & 1024 & V   & 36.8 & 39.7 & 34.4 & 31.2 & 41.3 & 36.4 & 33.3 & 34.7 & 34.0 \\
Molmo2-8B                   & 256  & V   & 32.0 & 40.5 & 37.7 & 31.2 & 31.7 & 34.6 & 36.0 & 44.0 & 40.0 \\
Video-XL-2-8B               & 256  & V   & 36.0 & 37.3 & 41.0 & 24.8 & 41.3 & 34.8 & 40.0 & 20.0 & 30.0 \\
\midrule
\rowcolor{bandgray} \multicolumn{12}{c}{\bf Agentic Video MLLMs} \\
EgoButler-GPT-4o$^\S$       & --   & T   & 34.4 & 42.1 & 29.5 & 30.4 & 44.4 & 36.2 & --   & --   & --   \\
EgoButler-Gemini 1.5 Pro$^\S$ & -- & T   & 36.0 & 37.3 & 45.9 & 30.4 & 34.9 & 36.9 & --   & --   & --   \\
Ego-R1-Qwen3.5-35B          & Full & V+T & 33.6 & 38.1 & 37.7 & 34.4 & 50.8 & 37.6 & 30.7 & 29.3 & 30.0 \\
AVP-Qwen3.5-35B             & Full & V+T & 29.6 & 32.5 & 27.9 & 28.0 & 33.3 & 30.2 & 32.0 & 25.3 & 28.7 \\
SiLVR-gpt-oss-120b          & Full & T   & 35.2 & 42.9 & 57.4 & 52.8 & 57.1 & 47.0 & 22.7 & 60.0 & 41.3 \\
EgoButler-gpt-oss-120b      & Full & T   & 40.0 & 40.5 & 47.5 & 45.6 & 46.0 & 43.2 & 29.3 & 37.3 & 33.3 \\
LightRAG$^\dagger$          & --   & --  & 40.8 & 48.4 & 67.2 & 50.4 & 44.4 & 48.8 & --   & --   & --   \\
HippoRAG$^\dagger$          & --   & --  & 48.8 & 60.3 & 70.5 & 60.8 & 66.7 & 59.6 & --   & --   & --   \\
Video-RAG$^\dagger$         & --   & --  & 49.6 & 56.3 & 67.2 & 55.2 & 54.0 & 55.4 & --   & --   & --   \\
EgoRAG$^\dagger$            & --   & --  & 40.0 & 56.3 & 62.3 & 54.4 & 52.4 & 52.0 & --   & --   & --   \\
Ego-R1-3B$^\dagger$         & --   & --  & 51.2 & 53.2 & 63.9 & 50.4 & 50.8 & 53.0 & --   & --   & --   \\
HippoMM$^\dagger$           & --   & --  & 45.6 & 53.2 & 70.5 & 55.2 & 58.7 & 54.6 & --   & --   & --   \\
M3-Agent$^\dagger$          & --   & --  & 44.4 & 54.8 & 62.3 & 56.8 & 54.0 & 53.5 & --   & --   & --   \\
EGAgent-Gemini 2.5 Pro$^\ddagger$ & 1\,FPS$\to$50 & V+T & 54.4 & 57.1 & 60.3 & 62.4 & \bf 74.6 & 57.5 & --   & --   & --   \\
WorldMM-GPT-5$^\dagger$     & Full & V+T & 62.4 & 64.3 & \bf 75.4 & 62.4 & \underline{71.4} & 65.6 & --   & --   & --   \\
WorldMM-Qwen3.5-35B$^*$     & Full & V+T & 49.6 & 54.8 & 54.1 & 62.4 & 60.3 & 56.0 & 48.0 & \underline{66.7} & 57.3 \\
MAGIC-Video-Qwen3.5-35B$^*$ & Full & V+T & \underline{67.2} & \underline{65.9} & 67.2 & \underline{66.4} & \bf 74.6 & \underline{67.6} & \underline{50.7} & \bf 78.7 & \underline{64.7} \\
\rowcolor{ourrow}
\name-Qwen3.5-35B (ours) & Full & V+T & \bf 70.4 & \bf 73.8 & \underline{73.8} & \bf 71.2 & \underline{71.4} & \bf 72.0 & \bf 64.0 & \bf 78.7 & \bf 71.3 \\
\bottomrule
\end{tabular}
\end{table}

\clearpage
\section{Qualitative Examples}
\label{app:rendered}

This appendix traces two EgoLifeQA questions through \name{} and the MAGIC-Video baseline, one through its frames and one through the two answering contexts, and shows the note that accompanies biographies that share a name.

\begin{figure}[!htb]
  \centering
  \includegraphics[width=1\textwidth]{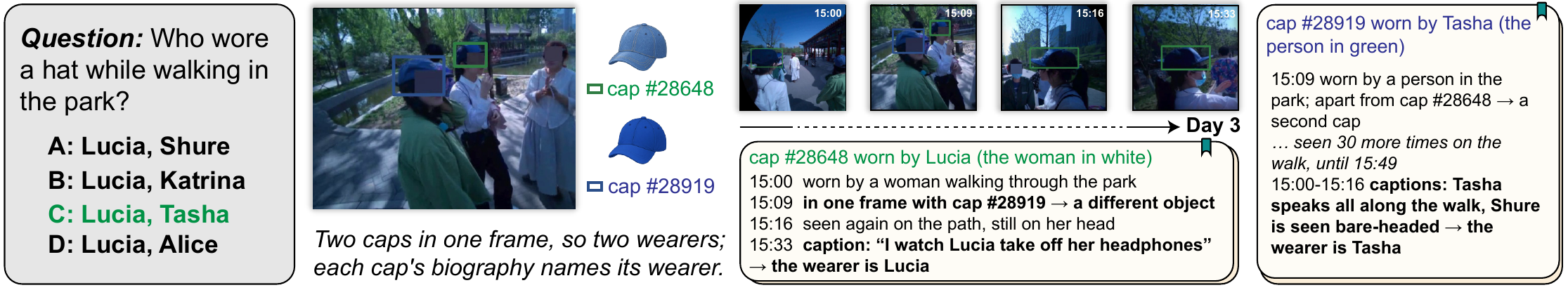}
  \caption{\textbf{Grounded biographies connect object identity to event context.}
    Two blue caps appear together at 15:09 on Day~3, so they are distinct instances despite sharing a description.
    \name{} keeps a biography for each cap and retrieves the surrounding episodes to identify its wearer: Lucia and Tasha (option~C).
    Cards summarize the retrieved biography and episodic evidence; colored boxes mark the two caps.}
  \label{fig:case}
\end{figure}

\paragraph{The hat question.}
Figure~\ref{fig:case} shows a question on which \name{} and the baseline diverge, \emph{who wore a hat while walking in the park}. The baseline's three searches return 28 episodes from three days, none of which says who wore a hat on the walk, and its answer is wrong. One \name{} search returns the person entity whose observation reads ``the woman with the blue hat'', the hat entity whose observation names Lucia, and a note that the two blue caps in the park are different objects. \name{} answers Lucia and Tasha (option C), which is right. The binding of hat to wearer, which the baseline's answer model had to infer, was established before the question was asked.

\definecolor{ctxgreen}{HTML}{2F7D43}
\definecolor{ctxcard}{HTML}{EEF4EE}
\definecolor{ctxbase}{HTML}{F4F4F2}
\definecolor{ctxred}{HTML}{B3261E}
\definecolor{ctxmuted}{HTML}{6D7266}
\definecolor{ctxink}{HTML}{232720}
\newcommand{\ctxkind}[1]{{\color{ctxink}\scriptsize\textbf{\texttt{#1}}}}
\newcommand{\ctxgut}[1]{{\color{ctxink}\textbf{#1}}}
\newcommand{\ctxtime}[1]{{\color{ctxmuted}#1}}
\newcommand{\ctxname}[1]{{\color{ctxink}\textbf{#1}}}
\newcommand{\ctxhi}[1]{{\color{ctxgreen}\textbf{#1}}}
\newcommand{\ctxbad}[1]{{\color{ctxred}\textbf{#1}}}
\newcommand{\ctxline}[1]{\par\vspace{2.5pt}\noindent#1}
\newcommand{\ctxsearch}[2]{\par\vspace{3pt}\noindent\textbf{Search #1}\enskip{\color{ctxmuted}#2}\par\vspace{1.5pt}}
\newcommand{\ctxone}[2]{\par\vspace{1.5pt}\noindent\makebox[84pt][l]{#1}\hspace{6pt}#2\par}
\newcommand{\ctxmore}[1]{\par\vspace{1pt}\noindent\makebox[84pt][l]{}\hspace{6pt}{\color{ctxmuted}[\ldots] #1}\par}
\newcommand{\ctxanswer}[1]{\par\vspace{3pt}\noindent\makebox[84pt][l]{\ctxgut{Answer}}\hspace{6pt}\textbf{#1}}
\newenvironment{ctxrows}{\noindent\begin{tabular}{@{}>{\raggedright\arraybackslash}p{84pt}@{\hspace{6pt}}>{\raggedright\arraybackslash}p{\dimexpr\linewidth-90pt\relax}@{}}}{\end{tabular}\par}
\newcommand{\ctxround}[1]{{\color{ctxmuted}\textcircled{\raisebox{0.4pt}{\tiny #1}}}}
\newcommand{\ctxitem}[2]{\par\vspace{2.5pt}\noindent#1\par\noindent\hspace*{7pt}\begin{minipage}[t]{\dimexpr\linewidth-7pt\relax}#2\end{minipage}\par}
\newcommand{\ctxpanel}[3]{\fcolorbox{#1}{#2}{\begin{minipage}{\dimexpr\textwidth-2\fboxsep-2\fboxrule\relax}#3\end{minipage}}}

\paragraph{Objects that share a name.}
When biographies that share a name are retrieved together, a closing note states which of them a shared frame proves to be different objects. In the hat question the two blue caps of Figure~\ref{fig:case} are seen together at 15:09 on Day~3, and the note reads:

\begin{center}\footnotesize\setlength{\fboxsep}{4pt}
\ctxpanel{ctxmuted}{ctxbase}{%
\ctxkind{[Retrieved entity]} Which same-name ids are the same object --- automatic tracking may have split one object into several ids. Each line below is a whole verdict:\par\noindent\hspace*{7pt}\begin{minipage}[t]{\dimexpr\linewidth-7pt\relax}
blue cap --- these are DIFFERENT objects: object\#28648, object\#28919 (none of them may be the same).
\end{minipage}}
\end{center}

The answer model's reasoning cites this line and gives the two caps two different wearers.

\paragraph{The scallion question.}
Figure~\ref{fig:qualitative} reproduces, abridged, the two answering contexts for EgoLifeQA question 54, asked on the first evening about the bundle of scallions in the wearer's hand.

\begin{figure}[!htbp]
\centering
\scriptsize
\setlength{\fboxsep}{3pt}
\ctxpanel{ctxmuted}{white}{%
\textbf{Question} (asked at DAY1 19:16:47).\enskip \emph{Who purchased the scallion in my hand?}\enskip (A) Me\enskip (B) Tasha\enskip (C) Alice\enskip (D) Lucia}\\[3pt]
\ctxpanel{ctxmuted}{ctxbase}{%
\textbf{MAGIC-Video-Qwen3.5-35B.} Five searches; 43 episodes, 10 triples, 64 frames.
\ctxsearch{1}{Who purchased the scallion I was holding?}
\begin{ctxrows}
\ctxkind{[Retrieved episode]} & \ctxtime{[DAY1 17:39:32 - DAY1 17:39:59]} I look left and right. I point at the large scallion in Alice's hand with my right hand and say, ``Take another look.'' [\ldots] Alice adds, ``We need it, but not that much.'' [\ldots]\\[1.5pt]
\ctxkind{[Retrieved episode]} & \ctxtime{[DAY1 18:25:32 - DAY1 18:26:00]} [\ldots] Lucia walks over holding a big scallion, watching and laughing the whole time. [\ldots] Lucia says, ``This is garlic.'' Alice wonders, ``Isn't this green onion?'' Lucia clarifies, ``This is garlic sprouts.'' [\ldots]\\
\end{ctxrows}
\ctxone{\ctxkind{[Retrieved semantic]}}{\ctxbad{(Lucia, holds, big scallion)}\quad (Alice, bought, garlic)}
\ctxmore{9 more episodes, 1 more triple}
\ctxsearch{2}{Who paid for the scallions or vegetables at the grocery store?}
\begin{ctxrows}
\ctxkind{[Retrieved episode]} & \ctxtime{[DAY1 17:36:36 - DAY1 17:37:00]} [\ldots] Alice says, ``This is 5 yuan.'' I ask, ``That one is 11 yuan, right?'' [\ldots] She then says, ``Let's buy this one.'' [\ldots]\\
\end{ctxrows}
\ctxmore{9 more episodes}
\ctxsearch{3}{Did I place the scallion in the shopping cart?}
\ctxone{\ctxkind{[Retrieved semantic]}}{(I, holds, shopping cart)\quad (I, pushes, shopping cart)\quad (I, shops with, Alice)}
\ctxmore{8 more episodes}
\ctxsearch{4}{Who paid for the groceries at the store checkout?}
\begin{ctxrows}
\ctxkind{[Retrieved episode]} & \ctxtime{[DAY1 18:17:30 - DAY1 18:17:59]} [\ldots] Lucia says, ``It's only filled here,'' and notes that everything is paid except one item that seems unpaid. Tasha adds that one item has been paid. [\ldots]\\
\end{ctxrows}
\ctxmore{9 more episodes, 4 more triples}
\ctxsearch{5}{Who paid for the scallions specifically or did I pay for the vegetables?}
\begin{ctxrows}
\ctxkind{[Retrieved episode]} & \ctxtime{[DAY1 18:17:06 - DAY1 18:17:30]} [\ldots] Shure hands me a very long bill, which I roll up with my hand. [\ldots] Tasha asks, ``Can it really be paid successfully?'' [\ldots]\\
\end{ctxrows}
\ctxmore{3 more episodes}
\ctxanswer{D} (Lucia) {\color{ctxred}\textbf{\texttimes}}}\\[3pt]
\ctxpanel{ctxgreen}{ctxcard}{%
\textbf{\name-Qwen3.5-35B.} Three searches; 10 episodes, 12 biographies, 64 frames.
\ctxsearch{1}{Who purchased the scallion that I was holding?}
\begin{ctxrows}
\ctxkind{[Retrieved entity]} & \ctxname{scallions in plastic packaging $\cdot$ object\#8606}\enskip \ctxtime{[DAY1 17:39:41 - DAY1 19:14:00]}\par
\ctxtime{[DAY1 17:39:41 - DAY1 17:39:52]} I held the bunch of scallions while examining them in the aisle. Alice and Tasha discussed if the item was necessary for their shopping. \ctxhi{I handed the scallions to Alice} at the end of the sequence.\par
\ctxtime{[DAY1 17:40:00 - DAY1 17:40:30]} \ctxhi{Alice held the pack of scallions} while discussing with Tasha whether to buy it. They debated the quantity needed for their plans and \ctxhi{eventually decided to keep it}.\par
\ctxtime{[DAY1 18:35:00 - DAY1 18:35:29]} I carried a bundle of scallions, then \ctxhi{Tasha took them from my bag to hold for me}. We discussed the items while walking with the group.\par
\ctxtime{also seen (no search selected these times yet, so no description here): DAY1 19:13:57 - DAY1 19:14:00}\\[1.5pt]
\ctxkind{[Retrieved episode]} & \ctxtime{[DAY1 18:35:01 - DAY1 18:35:29]} [\ldots] Tasha says, ``This scallion hit your foot,'' then adds, ``I'll hold it for you,'' and picks up the green onions from my bag. ``Oh, really? Thanks,'' I reply. [\ldots]\\
\end{ctxrows}
\ctxmore{4 more biographies, 4 more episodes}
\ctxsearch{2}{Who actually bought or paid for the scallions that I was holding?}
\begin{ctxrows}
\ctxkind{[Retrieved episode]} & \ctxtime{[DAY1 18:26:00 - DAY1 18:26:29]} I hold something and push the shopping cart over. [\ldots] Shure asks, ``Who cares, who brought the green onions?'' [\ldots] Alice confirms, ``Yes, it's garlic sprouts.'' [\ldots]\\
\end{ctxrows}
\ctxmore{2 more biographies, 1 more episode}
\ctxsearch{3}{Who paid for the scallions at the checkout?}
\begin{ctxrows}
\ctxkind{[Retrieved entity]} & \ctxname{check-out counter $\cdot$ object\#10148}\enskip \ctxtime{[DAY1 18:18:51 - DAY1 18:25:53]}\par
\ctxtime{[DAY1 18:25:03 - DAY1 18:25:53]} The group stood at the checkout counters to pay for their items. I used my phone to scan my cart. Tasha scanned a package of kitchen wipes. \ctxhi{Alice scanned a box of garlic sprouts.} [\ldots]\par
\ctxtime{also seen (no search selected these times yet, so no description here): DAY1 18:18:51 - DAY1 18:19:00, DAY1 18:23:45 - DAY1 18:23:49}\\
\end{ctxrows}
\ctxmore{4 more biographies, 3 more episodes}
\ctxanswer{C} (Alice) {\color{ctxgreen}\textbf{\checkmark}}}
\caption{\textbf{The biography of the bundle in the wearer's hand reaches its buyer, while the baseline's triple records only who held it.} Abridged verbatim from the two answering contexts. Bold green marks the phrases the answer turns on, red the line that misled the baseline, and each item stands under the search that returned it. The triple that names Lucia records who was holding the bundle at the checkout, not who chose and paid for it. The biography of the bundle in hand extends from the aisle to three minutes before the question and arrives with the first search, and the next two searches turn to the checkout, where the counter's observation shows Alice scanning it. The baseline's four further searches return prices, the cart and the bill, nothing that binds the bundle to a buyer.}
\label{fig:qualitative}
\end{figure}

\section{Prompts}
\label{app:prompts}

This section gives the prompts with which \name{} describes observations and searches the memory. Figures~\ref{fig:prompt-describe} and~\ref{fig:prompt-schema} give the description prompt on EgoLife, sent to the vision-language model once per observation with the crops and scene frames of Appendix~\ref{app:implementation}. On MultiHop-EgoQA the prompt keeps the same rules without the caption and transcript lines, so each observation is described from its own frames, and people are referred to by what is seen because the benchmark names no one. On the Test@Day stream the actors are game characters, and the prompt asks for the name the game gives each of them, since the questions refer to them by name. Test@Week uses the EgoLife memory unchanged. Figures~\ref{fig:prompt-controller} to~\ref{fig:prompt-controller-examples-2} give the controller prompt on EgoLife. At each round the controller reads the question and the round history and returns a JSON decision, either a search with its query or the answer. On the other benchmarks the controller prompt has the same structure, with the description of the recording, the time format and the examples written for it.

\begin{figure}[p]
\centering\setlength{\fboxsep}{5pt}
\fcolorbox{ctxmuted}{ctxbase}{\begin{minipage}{\dimexpr\textwidth-2\fboxsep-2\fboxrule\relax}\fontencoding{T1}\ttfamily\fontsize{6.4}{7.6}\selectfont\raggedright
These images come from MY OWN first-{}person (head-{}mounted) camera, and you observe ONE physical object in it. IMAGES 0-{}3 are close-{}up crops of the object. IMAGES 4-{}11 are full frames showing the SAME object in a RED BOUNDING BOX, in time order, so you can see what happens to it. They are evenly spaced about 1.5s apart, covering 10s in order, so they are one continuous stretch of time and IMAGE numbers run in that order.\par
\vspace{3pt}\par
[ME AND THE OTHER PEOPLE]\par
I am the camera wearer ({\color{ctxmuted}\textrm{\itshape\textless wearer\textquotesingle{}s name\textgreater}}), so I am always \symbol{34}I\symbol{34}/\symbol{34}my\symbol{34}/\symbol{34}me\symbol{34}, never \symbol{34}the camera wearer\symbol{34}, \symbol{34}the wearer\symbol{34} or \symbol{34}the user\symbol{34}. A hand or arm entering the frame edge with no body, torso or face attached is MINE; a hand you can trace back to a visible body is THAT person\textquotesingle{}s, however close it is to the boxed object. Credit other people\textquotesingle{}s actions to them, and name them by what you SEE — or by the NAME a line below gives.\par
\vspace{3pt}\par
[THE BOXED OBJECT]\par
Name what is INSIDE the red BOUNDING BOX this pipeline drew, never a container in the room itself. Read from the FULL frames which thing it encloses and how far it reaches; the crops show its detail, not its extent. It follows the OUTLINE of one thing, so whatever rests ON it, sits INSIDE it or stands BESIDE it is a DIFFERENT object, however central in the crop and however busy a hand is with it. So when it encloses a whole surface or piece of furniture with things on it, name the SURFACE or the FURNITURE and not any of those things: a box drawn around a table with a phone on it is the TABLE, even when the phone is what everyone is looking at. If it holds a person, they are an ACTOR: report what they DO and describe their look well enough to tell them from the others. A detector guessed \symbol{34}{\color{ctxmuted}\textrm{\itshape\textless detector label\textgreater}}\symbol{34}, a rough hint that may be wrong; trust the images.\par
\vspace{3pt}\par
[ACTIONS NEED EVIDENCE]\par
Claim a step only where you can cite it: a frame showing a hand IN CONTACT with the boxed object or the object DISPLACED against still surroundings, or a line below stating the step, cited by its [Ns] time. A hand near it, in front of it or busy with something else is not contact, and the camera moving is not the object moving. A line just BEFORE or after the frames counts too: an object comes into view once a hand has picked it up and leaves once it is put down, so nothing happening in the frames is not nothing happening to this object. With no frame and no line, abstain, even for something usually handled.\par
\vspace{3pt}\par
[SCENE NOTES AND DIALOGUE]\par
A narration log and a dialogue transcript cover this stretch, on the same clock as the frame times and running from before the frames to after them: the frames are only my pictures of this object ({\color{ctxmuted}\textrm{\itshape\textless time span of the observation\textgreater}}), and the record covers its story across the whole stretch. The * marks the lines overlapping the frames. My own lines are labelled \symbol{34}I:\symbol{34} and every other label is the person who spoke; the narration log is mine too, so a bare \symbol{34}I\symbol{34} in a line is me and never the boxed person, and a step credited to someone else stays theirs.\par
Narration log:\par
{\color{ctxmuted}\textrm{\itshape\textless caption lines within 40 s of the observation, as [Ns] text; * marks lines during its frames\textgreater}}\par
Dialogue transcript:\par
{\color{ctxmuted}\textrm{\itshape\textless transcript lines within 40 s of the observation, as [Ns] speaker: text\textgreater}}\par
\vspace{3pt}\par
[HOW TO USE THEM]\par
The IMAGES decide WHAT THE THING IS: a line names whatever its writer was attending to — usually the thing a HAND is busy with, not the thing the rectangle encloses — so a line may only make what you SEE more specific.\par
Use the narration log and the transcript to ENRICH what happened: the steps taken with this object and by whom, and what the people present said, asked and decided around it. Where no line concerns this object, write only what the frames show and keep it short.
\end{minipage}}
\caption{\textbf{Description prompt on EgoLife: instructions.} Sent once per observation with its crops and scene frames; the angle-bracketed parts are filled in per observation. The reply format follows in Figure~\ref{fig:prompt-schema}.}
\label{fig:prompt-describe}
\end{figure}

\begin{figure}[p]
\centering\setlength{\fboxsep}{5pt}
\fcolorbox{ctxmuted}{ctxbase}{\begin{minipage}{\dimexpr\textwidth-2\fboxsep-2\fboxrule\relax}\fontencoding{T1}\ttfamily\fontsize{6.4}{7.6}\selectfont\raggedright
[JSON SCHEMA]\par
Every field is read by someone who never sees these pictures and has no video to refer to, so name no recording of any kind — not \symbol{34}the video\symbol{34}, \symbol{34}the clip\symbol{34}, \symbol{34}the frames\symbol{34}, \symbol{34}this sequence\symbol{34}, \symbol{34}the footage\symbol{34}, \symbol{34}the session\symbol{34} — and use no IMAGE numbers outside the one field that asks for one: where you would write \symbol{34}throughout the video\symbol{34}, write plain past tense (\symbol{34}it stayed on the table\symbol{34}) or the stretch in words (\symbol{34}while we unpacked\symbol{34}). The red box is apparatus too, so no sentence starts with it or with \symbol{34}this object\symbol{34}: start with the THING, or with whoever acts on it. Return STRICT JSON:\par
\{\par
\leftskip=1.0em \symbol{34}object\_name\symbol{34}: \symbol{34}1-{}5 words: what it ACTUALLY is, read from the images (a \textbackslash{}\symbol{34}bottle\textbackslash{}\symbol{34} may be a \textbackslash{}\symbol{34}hand-{}soap dispenser\textbackslash{}\symbol{34}); \textbackslash{}\symbol{34}my \textless{}part\textgreater{}\textbackslash{}\symbol{34} if it is my own body or clothing. Never a person\textquotesingle{}s NAME — say what you SEE of them. Fall back to \textbackslash{}\symbol{34}{\color{ctxmuted}\textrm{\itshape\textless detector label\textgreater}}\textbackslash{}\symbol{34} only when the images are too unclear to tell\symbol{34},\par\leftskip=0pt\par
\leftskip=1.0em \symbol{34}is\_person\symbol{34}: \symbol{34}the JSON value true if the box is on a PERSON, false otherwise — true for a person only partly seen, false for part of my own body and for anything a person wears or carries\symbol{34},\par\leftskip=0pt\par
\leftskip=1.0em \symbol{34}appearance\symbol{34}: \symbol{34}\textless{}=15 words: color, material, what it is\symbol{34},\par\leftskip=0pt\par
\leftskip=1.0em \symbol{34}location\symbol{34}: \symbol{34}\textless{}=15 words: where this object BELONGS, which stays true between sightings even while a hand holds it now, e.g. the room or area it is in; \textbackslash{}\symbol{34}unclear\textbackslash{}\symbol{34} if you cannot tell.\symbol{34},\par\leftskip=0pt\par
\leftskip=1.0em \symbol{34}action\symbol{34}: \symbol{34}\textless{}=15 words: what is DONE to it over this stretch, first person if I do it; exactly \textbackslash{}\symbol{34}none\textbackslash{}\symbol{34} where you can cite nothing. FOR A PERSON: what THEY do\symbol{34},\par\leftskip=0pt\par
\leftskip=1.0em \symbol{34}action\_evidence\symbol{34}: \symbol{34}\textless{}=12 words: the IMAGE number showing the contact or displacement (\textbackslash{}\symbol{34}IMAGE 6: I grip the handle\textbackslash{}\symbol{34}), or the [Ns] time of the line stating the step; exactly \textbackslash{}\symbol{34}none\textbackslash{}\symbol{34} if action is \textbackslash{}\symbol{34}none\textbackslash{}\symbol{34}\symbol{34},\par\leftskip=0pt\par
\leftskip=1.0em \symbol{34}event\_desc\symbol{34}: \symbol{34}\textless{}=50 words: every step taken with this object in detail. exactly \textbackslash{}\symbol{34}no activity\textbackslash{}\symbol{34} if nothing happens.\symbol{34},\par\leftskip=0pt\par
\leftskip=1.0em \symbol{34}summary\symbol{34}: \symbol{34}1-{}3 sentences, retrievable by event-{}style questions: what happened with this object from beginning to end, who took part, and what was said or decided around it, enriched from the narration log and dialogue transcript. Write me as \textbackslash{}\symbol{34}I\textbackslash{}\symbol{34}. EVERY SENTENCE MUST BE ABOUT THIS OBJECT: one that would read the same with the box on anything else in the room does not belong, so drop it rather than reach three sentences — nobody touched it and no line named it is ONE sentence, and that is a complete answer. IF THE BOX HOLDS A PERSON, THE FIRST WORD IS THEM (\textbackslash{}\symbol{34}She ...\textbackslash{}\symbol{34}, \textbackslash{}\symbol{34}He ...\textbackslash{}\symbol{34}): the steps and the words are THEIRS, I appear only where they deal with me, and you refer to them by what you SEE (\textbackslash{}\symbol{34}the woman with pink hair\textbackslash{}\symbol{34}, \textbackslash{}\symbol{34}she\textbackslash{}\symbol{34}) — a name in a line is who SPOKE, so it can name the OTHER people in the story but never the boxed one.\symbol{34},\par\leftskip=0pt\par
\leftskip=1.0em \symbol{34}unique\_mark\symbol{34}: \symbol{34}one concrete mark telling THIS item from another of the same model — text, a logo, a sticker, a scratch, a stain; exactly \textbackslash{}\symbol{34}none\textbackslash{}\symbol{34} if there is none. FOR A PERSON: something they wear every day\symbol{34},\par\leftskip=0pt\par
\leftskip=1.0em \symbol{34}reid\_summary\symbol{34}: \symbol{34}1-{}3 sentences to recognise it hours later from another angle and in other light: only what holds in EVERY sighting, so no momentary action, angle, lighting or open/closed. Do not name another object of the same kind, and for a PERSON no names at all — cover hair, face and any permanent mark, glasses, build, age, clothing, as far as the images show\symbol{34},\par\leftskip=0pt\par
\leftskip=1.0em \symbol{34}source\_evidence\symbol{34}: \symbol{34}the [Ns] times of the lines you used, each with \textless{}=6 words on what matches; exactly \textbackslash{}\symbol{34}none\textbackslash{}\symbol{34} if you used none\symbol{34},\par\leftskip=0pt\par
\leftskip=1.0em \symbol{34}confidence\symbol{34}: \symbol{34}\textbackslash{}\symbol{34}low\textbackslash{}\symbol{34} when the box is too small or blurred to read, or does not hold one single thing throughout; \textbackslash{}\symbol{34}high\textbackslash{}\symbol{34} when it holds one thing and shows it clearly; \textbackslash{}\symbol{34}medium\textbackslash{}\symbol{34} between\symbol{34}\par\leftskip=0pt\par
\}
\end{minipage}}
\caption{\textbf{Description prompt on EgoLife: reply format}, continuing Figure~\ref{fig:prompt-describe}. The summary is the retrieval text of the observation.}
\label{fig:prompt-schema}
\end{figure}

\begin{figure}[p]
\centering\setlength{\fboxsep}{5pt}
\fcolorbox{ctxmuted}{ctxbase}{\begin{minipage}{\dimexpr\textwidth-2\fboxsep-2\fboxrule\relax}\fontencoding{T1}\ttfamily\fontsize{6.4}{7.6}\selectfont\raggedright
You are a reasoning agent for a multimodal video memory retrieval system.\par
Your job is to decide whether to stop and answer, or to search memory for more evidence.\par
\vspace{3pt}\par
\# Decision Modes:\par
1. **search**: Retrieve memory to begin, continue, or extend progress toward the answer.\par
\leftskip=1.5em -{} Write the search query as a **natural-{}language sentence or question** (NOT a list of keywords).\par\leftskip=0pt\par
\leftskip=2.5em Good:  \symbol{34}Who handed the black marker to Shure?\symbol{34}\par\leftskip=0pt\par
\leftskip=2.5em Bad:   \symbol{34}black marker Shure hand location\symbol{34}\par\leftskip=0pt\par
\leftskip=1.5em -{} The retrieval system uses semantic embedding similarity, so natural sentences work much better than keyword lists.\par\leftskip=0pt\par
\leftskip=1.5em -{} Each round, try a **different angle** — do not rephrase the same query.\par\leftskip=0pt\par
2. **answer**: Stop searching because the accumulated results are sufficient.\par
\leftskip=1.5em -{} If 2+ consecutive rounds returned \symbol{34}[No new results]\symbol{34}, you MUST answer with what you have.\par\leftskip=0pt\par
\vspace{3pt}\par
\# Context Inputs:\par
-{} Current Query\par
-{} Round History: Log of past retrieval rounds. Each round is written in this format:\par
\vspace{3pt}\par
\leftskip=1.0em \#\#\# Round N\par\leftskip=0pt\par
\leftskip=1.0em Decision: \textless{}search|answer\textgreater{}\par\leftskip=0pt\par
\leftskip=1.0em Search Query: \textless{}query text\textgreater{}\par\leftskip=0pt\par
\leftskip=1.0em Retrieved:\par\leftskip=0pt\par
\leftskip=1.0em \textless{}retrieved items summary\textgreater{}\par\leftskip=0pt\par
\vspace{3pt}\par
\# STRICT OUTPUT RULES:\par
-{} Always decide **first**: \symbol{34}search\symbol{34} or \symbol{34}answer\symbol{34}.\par
-{} If decision = \symbol{34}search\symbol{34}: Must include \symbol{34}search\_query\symbol{34} (a single concise query string).\par
-{} If decision = \symbol{34}answer\symbol{34}: Do NOT include \symbol{34}search\_query\symbol{34}.\par
-{} Always output valid JSON only, no extra commentary.\par
\vspace{3pt}\par
\# Output Format:\par
\{\par
\leftskip=1.0em \symbol{34}decision\symbol{34}: \symbol{34}search\symbol{34} | \symbol{34}answer\symbol{34},\par\leftskip=0pt\par
\leftskip=1.0em \symbol{34}search\_query\symbol{34}: \symbol{34}\textless{}str\textgreater{}\symbol{34}\par\leftskip=0pt\par
\}\par
\vspace{3pt}\par
\# Object timelines\par
-{} A round\textquotesingle{}s results come in up to two parts. \symbol{34}Moments retrieved\symbol{34} are excerpts of the recording, in time order. If an \symbol{34}Objects now known\symbol{34} part follows, each entry there is ONE tracked object and the times it was seen. A time a search has covered carries a description; the remaining times are listed without one. The ids come from automatic tracking and re-{}identification, which is imperfect both ways: one object may be split into two ids, and two objects may be merged under one id. Appearances can also be missed, so a timeline is what was detected, not a complete history.\par
-{} A \symbol{34}Which same-{}name ids are the same object\symbol{34} list after the entries sorts out the ids that share a name. A \symbol{34}these may be THE SAME object\symbol{34} line means every id on it could be the same single thing, so their times may all belong to one object; a \symbol{34}these are DIFFERENT objects\symbol{34} line means those ids were seen apart at the same moment, so they cannot be merged. An id can appear on more than one may-{}be line — it may be either of them, while those two are not each other. So before concluding \symbol{34}the same X\symbol{34}, or counting how many Xs there were, read those lines.\par
-{} When the question asks **how many times** something happened, **which came first**, or **who did it first**, search for the object AND the interaction (\symbol{34}Who picked up the screwdriver?\symbol{34}), not just the object\textquotesingle{}s name: an object\textquotesingle{}s timeline lists the times it was seen, which is exactly the axis such a question needs.\par
-{} **To fill in a listed time, search by time**: a query naming a time — with its DAY, like \symbol{34}What was I doing between DAY1 12:30:44 and DAY1 12:41:00?\symbol{34} — reads out the recording at that time INSTEAD of searching. It can only return what is already there, so use it for a time you already have (for example a time listed for an object but not yet described), one time or range per query. Always write the DAY with the clock time: without it the day has to be guessed. A query that only bounds time on one side (\symbol{34}before DAY1 17:20\symbol{34}, \symbol{34}after DAY1 20:32\symbol{34}) is not a readout — it searches by meaning as usual.\par
-{} A time listed for an object with no description means no search has covered that time yet — one search by time can fill it in. If that search returns nothing, answer from the times themselves.\par
-{} An entry\textquotesingle{}s description was written for that ONE object, so it can attach what was said or done nearby to the wrong object. Where an entry conflicts with a \symbol{34}Moments retrieved\symbol{34} excerpt, trust the excerpt, and search for that moment to see it in full.
\end{minipage}}
\caption{\textbf{Controller prompt on EgoLife: instructions.} The system prompt of the controller; the question and the round history follow it as the user message. The few-shot examples follow in Figure~\ref{fig:prompt-controller-examples-1}.}
\label{fig:prompt-controller}
\end{figure}

\begin{figure}[p]
\centering\setlength{\fboxsep}{5pt}
\fcolorbox{ctxmuted}{ctxbase}{\begin{minipage}{\dimexpr\textwidth-2\fboxsep-2\fboxrule\relax}\fontencoding{T1}\ttfamily\fontsize{6.4}{7.6}\selectfont\raggedright
\# Few-{}shot Examples:\par
\#\# Example 1\par
Query: Who gives the graduation gift to Maria?\par
Round History: []\par
\vspace{3pt}\par
\#\#\# Response:\par
\{\par
\leftskip=1.0em \symbol{34}decision\symbol{34}: \symbol{34}search\symbol{34},\par\leftskip=0pt\par
\leftskip=1.0em \symbol{34}search\_query\symbol{34}: \symbol{34}Who gave a graduation gift to Maria?\symbol{34}\par\leftskip=0pt\par
\}\par
\vspace{3pt}\par
\#\# Example 2\par
Query: Who gives the graduation gift to Maria?\par
Round History:\par
\#\#\# Round 1\par
Decision: search\par
Search Query: Who gave a graduation gift to Maria?\par
Retrieved:\par
[DAY1 10:30:00 -{} DAY1 10:31:30] (30sec)\par
I watch Luis hand a wrapped gift to Maria at the ceremony.\par
(Luis, gives, graduation gift to Maria)\par
\vspace{3pt}\par
\#\#\# Response:\par
\{\par
\leftskip=1.0em \symbol{34}decision\symbol{34}: \symbol{34}search\symbol{34},\par\leftskip=0pt\par
\leftskip=1.0em \symbol{34}search\_query\symbol{34}: \symbol{34}What is Luis\textquotesingle{}s relationship to Maria?\symbol{34}\par\leftskip=0pt\par
\}\par
\vspace{3pt}\par
\#\# Example 3\par
Query: Who gives the graduation gift to Maria?\par
Round History:\par
\#\#\# Round 1\par
Decision: search\par
Search Query: Who gave a graduation gift to Maria?\par
Retrieved:\par
[DAY1 10:30:00 -{} DAY1 10:31:30] (30sec)\par
I watch Luis hand a wrapped gift to Maria at the ceremony.\par
\vspace{3pt}\par
\#\#\# Round 2\par
Decision: search\par
Search Query: What is Luis\textquotesingle{}s relationship to Maria?\par
Retrieved:\par
(Luis, is brother of, Maria)\par
\vspace{3pt}\par
\#\#\# Response:\par
\{\par
\leftskip=1.0em \symbol{34}decision\symbol{34}: \symbol{34}answer\symbol{34}\par\leftskip=0pt\par
\}\par
\vspace{3pt}\par
\#\# Example 4 (incorporate discovered clues into later queries)\par
Query: Who used the microwave last on the first floor?\par
Round History:\par
\#\#\# Round 1\par
Decision: search\par
Search Query: Who used the microwave on the first floor?\par
Retrieved:\par
[DAY1 20:32:00 -{} DAY1 20:32:30] (30sec)\par
Lucia asks, \symbol{34}Can it fit?\symbol{34} I reply, \symbol{34}Yes.\symbol{34} I put a plate in the microwave.\par
(Lucia, adjusts, microwave)\par
(I, uses, microwave)\par
\vspace{3pt}\par
\#\#\# Response:\par
\{\par
\leftskip=1.0em \symbol{34}decision\symbol{34}: \symbol{34}search\symbol{34},\par\leftskip=0pt\par
\leftskip=1.0em \symbol{34}search\_query\symbol{34}: \symbol{34}Did Lucia or anyone else use the first-{}floor microwave after DAY1 20:32?\symbol{34}\par\leftskip=0pt\par
\}\par
\vspace{3pt}\par
\#\# Example 5 (no new results — stop early)\par
Query: Where did I put the red box?\par
Round History:\par
\#\#\# Round 1\par
Decision: search\par
Search Query: Where did I place the red box?\par
Retrieved:\par
[No new results]\par
\vspace{3pt}\par
\#\#\# Round 2\par
Decision: search\par
Search Query: What happened with the red box recently?\par
Retrieved:\par
[No new results]\par
\vspace{3pt}\par
\#\#\# Response:\par
\{\par
\leftskip=1.0em \symbol{34}decision\symbol{34}: \symbol{34}answer\symbol{34}\par\leftskip=0pt\par
\}
\end{minipage}}
\caption{\textbf{Controller prompt on EgoLife: few-shot examples 1--5}, continuing Figure~\ref{fig:prompt-controller}.}
\label{fig:prompt-controller-examples-1}
\end{figure}

\begin{figure}[p]
\centering\setlength{\fboxsep}{5pt}
\fcolorbox{ctxmuted}{ctxbase}{\begin{minipage}{\dimexpr\textwidth-2\fboxsep-2\fboxrule\relax}\fontencoding{T1}\ttfamily\fontsize{6.4}{7.6}\selectfont\raggedright
\#\# Example 6 (an object timeline carries the counting axis — query the interaction)\par
Query: How many times did I use the frying pan today?\par
Round History:\par
\#\#\# Round 1\par
Decision: search\par
Search Query: When did I cook with the frying pan?\par
Retrieved:\par
Moments retrieved (in time order):\par
[DAY1 09:02:00 -{} DAY1 09:02:30] (30sec)\par
I crack two eggs into the frying pan on the front burner.\par
Objects now known (each is one tracked object and the times it was seen, not a moment in the video — you may search any of these times):\par
[DAY1 09:02:10 -{} DAY1 19:45:31] (frying pan · object\#210) visible at these times:\par
\leftskip=1.0em [DAY1 09:02:10 -{} DAY1 09:14:02] I fry eggs in the pan on the front burner.\par\leftskip=0pt\par
\leftskip=1.0em also seen (no search selected these times yet, so no description here): DAY1 12:30:44 -{} DAY1 12:41:00, DAY1 19:40:12 -{} DAY1 19:45:31\par\leftskip=0pt\par
\vspace{3pt}\par
\#\#\# Response:\par
\{\par
\leftskip=1.0em \symbol{34}decision\symbol{34}: \symbol{34}search\symbol{34},\par\leftskip=0pt\par
\leftskip=1.0em \symbol{34}search\_query\symbol{34}: \symbol{34}Was I cooking with the frying pan around midday, or was it just sitting on the stove?\symbol{34}\par\leftskip=0pt\par
\}\par
\vspace{3pt}\par
\#\# Example 7 (fill in one listed time — search by time, DAY and both ends)\par
Query: When did I put the paintbrush down?\par
Round History:\par
\#\#\# Round 1\par
Decision: search\par
Search Query: When was I holding the paintbrush?\par
Retrieved:\par
Objects now known (each is one tracked object and the times it was seen, not a moment in the video — you may search any of these times):\par
[DAY2 15:02:11 -{} DAY2 15:40:03] (wooden paintbrush · object\#87) visible at these times:\par
\leftskip=1.0em [DAY2 15:02:11 -{} DAY2 15:09:46] I hold a wooden paintbrush and load it with paint from the tin.\par\leftskip=0pt\par
\leftskip=1.0em also seen (no search selected these times yet, so no description here): DAY2 15:38:20 -{} DAY2 15:40:03\par\leftskip=0pt\par
\vspace{3pt}\par
\#\#\# Response:\par
\{\par
\leftskip=1.0em \symbol{34}decision\symbol{34}: \symbol{34}search\symbol{34},\par\leftskip=0pt\par
\leftskip=1.0em \symbol{34}search\_query\symbol{34}: \symbol{34}What was I doing between DAY2 15:38:20 and DAY2 15:40:03?\symbol{34}\par\leftskip=0pt\par
\}\par
\vspace{3pt}\par
\#\# Example 8 (same-{}named objects — check the identity list before counting)\par
Query: Who used the screwdriver first?\par
Round History:\par
\#\#\# Round 1\par
Decision: search\par
Search Query: Who picked up the screwdriver?\par
Retrieved:\par
Objects now known (each is one tracked object and the times it was seen, not a moment in the video — you may search any of these times):\par
[DAY1 11:20:04 -{} DAY1 11:20:14] (screwdriver · object\#1232) visible at these times:\par
\leftskip=1.0em [DAY1 11:20:04 -{} DAY1 11:20:14] A screwdriver with a black handle rests on the table.\par\leftskip=0pt\par
[DAY1 11:20:11 -{} DAY1 11:20:18] (screwdriver · object\#1242) visible at these times:\par
\leftskip=1.0em [DAY1 11:20:11 -{} DAY1 11:20:18] The screwdriver remains stationary on the table.\par\leftskip=0pt\par
[DAY1 14:02:30 -{} DAY1 14:02:41] (screwdriver · object\#1300) visible at these times:\par
\leftskip=1.0em [DAY1 14:02:30 -{} DAY1 14:02:41] Shure turns a screw with a screwdriver.\par\leftskip=0pt\par
Which same-{}name ids are the same object — automatic tracking may have split one object into several ids. Each line below is a whole verdict:\par
\leftskip=1.0em screwdriver — these may be THE SAME object: object\#1232, object\#1242.\par\leftskip=0pt\par
\leftskip=1.0em screwdriver — these may be THE SAME object: object\#1242, object\#1300.\par\leftskip=0pt\par
\leftskip=1.0em screwdriver — these are DIFFERENT objects: object\#1232, object\#1300.\par\leftskip=0pt\par
\vspace{3pt}\par
\#\#\# Response:\par
\{\par
\leftskip=1.0em \symbol{34}decision\symbol{34}: \symbol{34}search\symbol{34},\par\leftskip=0pt\par
\leftskip=1.0em \symbol{34}search\_query\symbol{34}: \symbol{34}Who was holding or turning a screwdriver while we assembled things?\symbol{34}\par\leftskip=0pt\par
\}
\end{minipage}}
\caption{\textbf{Controller prompt on EgoLife: few-shot examples 6--8}, continuing Figure~\ref{fig:prompt-controller-examples-1}.}
\label{fig:prompt-controller-examples-2}
\end{figure}

\section{Limitations}
\label{app:limitations}
\name{} estimates instance identity from visual evidence, and reliable tracking and re-identification across long videos remain open challenges. Association errors may assign an observation to the wrong biography, while contextual descriptions may attribute a nearby action to the wrong entity. Retrieval over the larger memory graph also incurs higher latency than caption-only retrieval (Appendix~\ref{app:latency}). These limitations suggest concrete directions for further improving association reliability, action attribution, and retrieval efficiency within the proposed framework.

\end{document}